\documentclass[sigconf, screen, nonacm]{acmart}

\usepackage{bm}
\usepackage{pifont}
\usepackage{stfloats}
\newcommand{\cmark}{\ding{51}}%
\newcommand{\xmark}{\ding{55}}%

\AtBeginDocument{%
  }

\setcopyright{acmlicensed}
\copyrightyear{2024}
\acmYear{2024}
\acmDOI{XXXXXXX.XXXXXXX}

\acmConference[SIGSPATIAL 2024]{32nd ACM SIGSPATIAL International Conference on Advances in Geographic Information Systems (ACM SIGSPATIAL 2024)}{October 29 - November 1, 2024}{Atlanta, Georgia, USA}
\acmISBN{978-1-4503-XXXX-X/18/06}

\acmSubmissionID{126}

\begin{document}
\title[GFM4MPM: Towards Geospatial Foundation Models for Mineral Prospectivity Mapping]{GFM4MPM: Towards Geospatial Foundation Models for \\ Mineral Prospectivity Mapping\\ \vspace{0.1in}\large{(Manuscript is under review)}}


\author{Angel Daruna}
\authornote{Both authors contributed equally to this research.}
\email{angel.daruna@sri.com}
\orcid{0000-0002-9620-036X}
\affiliation{%
  \institution{SRI International}
  \city{Princeton}
  \state{NJ}
  \country{USA}
}

\author{Vasily Zadorozhnyy}
\authornotemark[1]
\email{vasily.zadorozhnyy@sri.com}
\orcid{0000-0002-0379-7348}
\affiliation{%
  \institution{SRI International}
  \city{Princeton}
  \state{NJ}
  \country{USA}
}

\author{Georgina Lukoczki}
\email{gina.lukoczki@uky.edu}
\orcid{0000-0002-0661-0198}
\affiliation{%
 \institution{University of Kentucky}
 \city{Lexington}
 \state{KY}
 \country{USA}
 }

\author{Han-Pang Chiu}
\email{han-pang.chiu@sri.com}
\orcid{0000-0001-8621-3725}
\affiliation{%
  \institution{SRI International}
  \city{Princeton}
  \state{NJ}
  \country{USA}
}


\renewcommand{\shortauthors}{Daruna et al.}

\begin{abstract}
Machine Learning (ML) for Mineral Prospectivity Mapping (MPM) remains a challenging problem as it requires the analysis of associations between large-scale multi-modal geospatial data and few historical mineral commodity observations (positive labels). Recent MPM works have explored Deep Learning (DL) as a modeling tool with more representation capacity. However, these overparameterized methods may be more prone to overfitting due to their reliance on scarce labeled data. While a large quantity of \textit{unlabeled} geospatial data exists, no prior MPM works have considered using such information in a self-supervised manner. Our MPM approach uses a masked image modeling framework to pretrain a backbone neural network in a self-supervised manner using unlabeled geospatial data alone. After pretraining, the backbone network provides feature extraction for downstream MPM tasks. We evaluated our approach alongside existing methods to assess mineral prospectivity of Mississippi Valley Type (MVT) and Clastic-Dominated (CD) Lead-Zinc deposits in North America and Australia. Our results demonstrate that self-supervision promotes robustness in learned features, improving prospectivity predictions. Additionally, we leverage explainable artificial intelligence techniques to demonstrate that individual predictions can be interpreted from a geological perspective.
\end{abstract}




\maketitle

\vspace{-0.25cm}
\section{Introduction}\label{s:introduction}

Data-driven mineral prospectivity mapping (MPM) seeks to apply Machine Learning (ML) techniques for improving traditional mineral commodity assessment, which is a time-consuming manual process. Minerals are naturally occurring, inorganic solid substances with specific chemical compositions and crystalline structure formed through geological processes in Earth's crust. Monitoring the supply of scarce mineral commodities (e.g. critical minerals), that are essential for the economy and national security of a country, is an ongoing concern in the real world~\cite{emsbo2021geological}.

ML MPM, which is the application of our interest, aims to estimate the prospectivity (i.e., presence likelihood) of an exploration target, such as a mineral deposit, at new locations based on knowledge from existing locations. MPM is typically approached as a binary classification problem. Key factors for mineral deposit formation are identified and converted into raster data, that serve as the input features for this classification problem. These key factors are mappable proxies correlated with the mineral deposit formation. Historical records of mineral deposit locations can be mapped using georeferencing techniques to correspond with the input features, providing labels for this classification problem. ML MPM then learns a function that predicts the presence of a mineral deposit given the explanatory input features (i.e., raster data).

\begin{figure*}
  \includegraphics[width=0.9\textwidth]{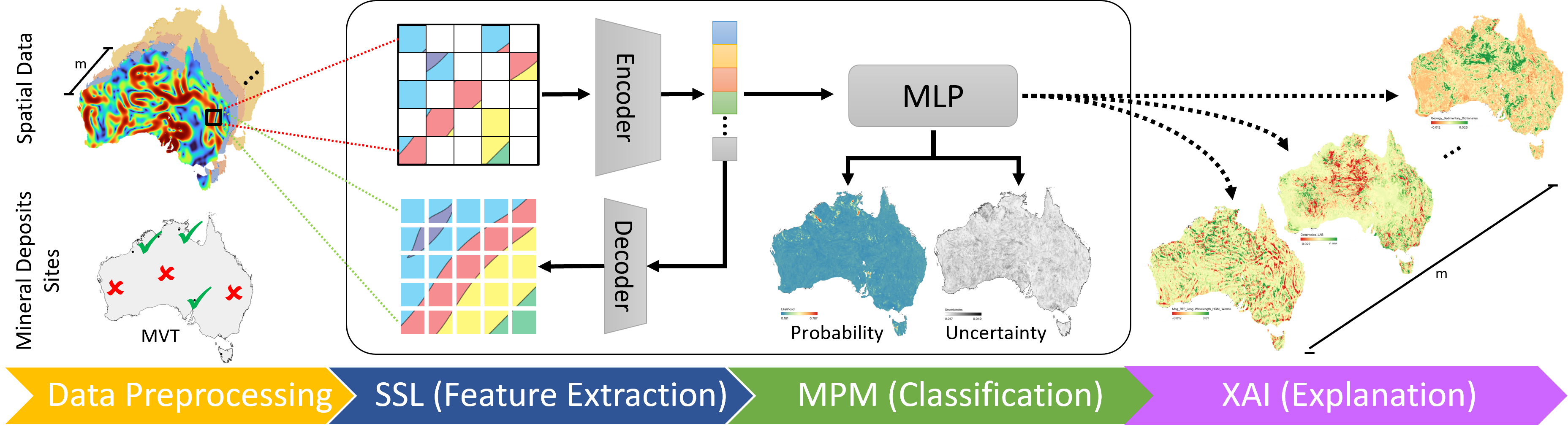}
  \vspace{-0.25cm}
  \caption{Overview of our Mineral Prospectivity Mapping (MPM) approach. Left-to-right: geospatial data preprocessing; geospatial \underline{S}elf-\underline{S}upervised \underline{L}earning; MPM using extracted SSL features; and validating MPM predictions using feature importances} 
  \Description{}
  \label{fig:overview}
  \vspace{-0.25cm}
\end{figure*}

ML MPM remains a challenging problem, because typically few historical mineral commodity records are available to learn from large-scale multi-modal geospatial data. Recent works~\cite{sun2020data,yin2023mineral} have recognized the potential of Deep Learning (DL) as an advanced ML tool for MPM, as DL is capable of finding more complex geospaial relations than traditional ML methods~\cite{zuo2020geodata}. However, there are still significant difficulties in exploiting DL for MPM. First, mineral deposit formation (i.e., geological mineralization) is extremely rare. The rarity of mineralization leads to label scarcity, which increases the risk of overfitting due to the over-parametrization of DL architectures. Second, DL predictions can be challenging to be interpret and validate from a geological point of view. The complex nature of these techniques obfuscates how input features affect output predictions. Third, incomplete domain knowledge and errors exist in our understanding of mineralization and how we operationalize MPM~\cite{zuo2021uncertainties}. Therefore, uncertainty should also be estimated and presented in predictions from DL methods.

We propose a new unified framework that addresses these DL difficulties for MPM, using self-supervised learning, explainable artificial intelligence, and epistemic uncertainty modeling shown in Figure~\ref{fig:overview}. Our approach to the MPM problem presents a new direction in which a single backbone model is pretrained, then re-used for various downstream MPM tasks (i.e., prediction of a mineral deposit type is a single task). Training a single backbone model that can be used across multiple tasks (multiple mineral deposit types) is made possible through a masked image modeling objective that exclusively uses \text{unlabeled} data. In this way, the learned features are a generic representation of the localized geospatial information that can be used for MPM. In addition to self-supervision, we treat MPM as an imbalanced, \textit{positive-unlabeled} binary classification problem. This problem formulation leads to a novel undersampling technique, that reduces the chance of labeling unknown true positives as negative samples in previous works. We also leverage the integrated gradients feature attribution method~\cite{integrad_mukun17} to offer insights into which explanatory features most influenced each MPM prediction. Our method produces predictions that pair each presence likelihood with an epistemic uncertainty measure, based on the Monte-Carlo Dropout~\cite{mcdropout_gal16} method. Our capabilities in providing feature attributions and uncertainties reduce the barriers to validating the DL MPM predictions from a geological perspective.

We validated our DL MPM method, by recreating two large-scale mineral assessment datasets. We followed the experimental settings of~\cite{LAWLEY2022104635} to recreate datasets for mineral assessments of Mississippi Valley-Type (MVT) and Clastic-Dominated (CD) Lead-Zinc deposits in North America and Australia. We compared our performance with five baselines based on prior MPM works~\cite{agterberg1989systematic, LAWLEY2022104635, brown2000artificial, sun2020data, yin2023mineral}. We also use a suite of six classification metrics to confirm consistency in results. Results in all experiments are reported as the average and standard deviation of repeated trials, which are conducted with different training data splits. To better contextualize the prediction results, we report an analysis of each model's computational complexity. Additional ablation studies were performed to substantiate our design choices, including analysis of self-supervision and our new undersampling technique.

Overall, our results indicate that self-supervised learning promotes feature robustness and model generalization. We believe our work presents an important step towards the new direction of building geospatial foundation models for mineral exploration. Our contributions are as follows:

\begin{enumerate}
    \item We propose a self-supervised MPM pretraining method based on masked image modeling that leverages unlabeled geospatial data.
    \item We incorporate explainable AI and epistemic uncertainty modeling into our DL MPM method to validate individual DL MPM predictions from a geological perspective.
    \item We treat MPM as an imbalanced, positive-unlabaled binary classification problem, leading to the formulation of a novel undersampling method for MPM.
    \item We validate our approach with comprehensive experiments and ablation studies, demonstrating our improved prediction performance and robustness over baseline methods.
\end{enumerate}

\section{Background \& Related Work}\label{s:problem_statement}

\subsection{Preliminaries}\label{ss:preliminaries}
We model geospatial information (e.g., geophysics, geology, geochemistry) using a multi-band georeferenced raster framework. A raster is a representation of a two dimensional plane as a matrix of $n$ pixels organized into a rectangular grid of $r$ rows and $c$ columns where each pixel contains information. A \textit{georeferenced} raster typically encodes geospatial information with a projection of the raster onto the Earth's surface using an Earth coordinate reference system. Therefore, every pixel on a georeferenced raster has a corresponding area on Earth's surface. Our multi-band georeferenced raster framework contains $m$ rasters that represent geospatial information. We denote a multi-band georeferenced raster as $X=\{x_1,\ldots,x_n\}$ where $x_i \in \mathbb{R}^{m\times 1}$. Here, the multi-band raster $X \in \mathbb{R}^{m \times n}$, or equivalently in raster format, $X \in \mathbb{R}^{m \times r \times c}$.

Mineral exploration is often approached as a binary classification problem~\cite{zuo2020geodata} in which the goal is to estimate the likelihood for the presence of some mineral exploration target (e.g., mineral deposit). In the multi-band georeferenced raster framework, a single-band georeferenced raster can be used to encode the exploration target presence in each pixel, denoted as $Y=\{y_1,\ldots,y_n\}$. Therefore, the georeferenced rasters represent both the explanatory features (i.e., $X$) and target features (i.e., $Y$) for mineral exploration.

\subsection{Problem Statement}\label{ss:problem_defn}

MPM seeks to estimate the prospectivity (i.e., presence likelihood) of an exploration target at new locations based on knowledge from existing locations. More formally, the inputs to our problem include a multi-band georeferenced raster $X$ and a history of observations where the exploration target is present to derive our labels $Y$ (e.g., mineral deposit). The goal of MPM is to find a function $\mathcal{F}$ of the localized geospatial data in a location $x'_i \in X$ that estimates its presence likelihood $p(y_i=\textrm{present} | x'_i)$, i.e., $\mathcal{F}(x'_i) \approx p(y_i=\textrm{present} | x'_i)$. In ML MPM, $\mathcal{F}$ can take many forms, see Section~\ref{ss:related}, but typically $\mathcal{F}$ is found through some procedure that optimizes a supervised objective function $\mathcal{L}_S$ of the available labels $y_i$ with corresponding inputs $x'_i$, i.e., $\mathcal{L}_S\big(\mathcal{F}(x'_i), y_i\big)$. The optimization procedure for $\mathcal{F}$ incrementally updates the function to find one with the best objective value $\mathcal{L}$, providing our problem solution.

To illustrate, consider the case of mapping mineral deposit presence likelihood. The multi-band georeferenced raster contains geospatial explanatory features for the mineral deposit presence. These explanatory feature layers can include geophysics, geochemistry, geology, and other types of proxies correlated with the mineral deposit's presence. Data for these features comes from a variety of sources and in many formats (e.g., maps of lithology, geochemical samples, magnetic anomaly measurements), but all source data can be georeferenced and rasterized to be represented as $X$. Additionally, all records of mineral deposits can be converted into a single-band georeferenced raster to be represented as $Y$. Every pixel within the single-band georeferenced raster that overlaps with a historical observation can be labeled present while remaining pixels are unknown. As a simplifying assumption, prior MPM work typically labels these unknown pixels as absent. With $X$ and $Y$ realized, some flavor of ML MPM can be applied to define the function $\mathcal{F}$, objective $\mathcal{L}$, and the optimization procedure. For example, in~\cite{brown2000artificial} the Levenberg-Marquardt algorithm, which has a sum of residuals objective function, was used to optimize the parameters of a multi-layer perceptron, a form of artificial neural network (i.e., $\mathcal{F}$).

\subsection{Related Work}\label{ss:related}

Traditional MPM methods, such as Weights of Evidence (WoE)~\cite{agterberg1989systematic} and Fuzzy Logic~\cite{an1991application}, are popular for their ease of interpretation and implementation. For example, WoE applies conditional probability theory to assess the associations between explanatory features and known deposits. ML MPM methods have been considered to learn more complex associations between explanatory variables and deposit sites, providing better predictions over traditional methods. Prior ML MPM works have used Artificial Neural Networks (ANN)~\cite{singer1996application,brown2000artificial}, Support Vector Machines~\cite{zuo2011support}, Random Forests (RF)~\cite{rodriguez2014predictive}, and Gradient Boosting~\cite{LAWLEY2022104635}. Decision Tree-based methods (e.g., RF, GBM) have also been popular for MPM. They have the modeling capacity to represent non-linear functions, and can assess the importance of explanatory features across all predictions.

Most recently, the potential of Deep Learning (DL) for MPM has been of interest due to its dominance in many other fields. In~\cite{sun2020data}, Convolutional Neural Networks (CNN) have been used and found to outperform the RF method. In~\cite{yin2023mineral}, transformer-based architectures were explored and found to outperform ANN. These works demonstrate how DL MPM can extract new and unknown patterns that may not be identified by prior models. However, DL architectures tend to be over-parameterized, which might increase the risk of overfitting $\mathcal{F}$ to training data in situations with few known mineralizations (i.e., label scarcity)~\cite{zuo2020geodata}. DL MPM predictions are also difficult to be interpreted from a geological point of view due to the complex nature of these techniques. Last, our understanding of mineralization and how we operationalize MPM for ML is inherently uncertain~\cite{zuo2021uncertainties}. While such uncertainties should be represented in MPM, they are rarely considered in DL or ML MPM predictions. Our work seeks to address these issues, by using self-supervised learning, explainable artificial intelligence, and epistemic uncertainty modeling, respectively. More details are described in Section~\ref{s:approach}.
 
\section{Approach}\label{s:approach}

Our approach to the MPM problem presents a new direction in which a single backbone model is pretrained, then re-used for various downstream MPM tasks (e.g., across mineral deposit types). Pretraining a model that can be used across multiple mineral deposit types is made possible through a self-supervised objective that exclusively uses unlabeled data. We treat MPM as an imbalanced, \textit{positive-unlabaled} binary classification problem that leads to a novel undersampling technique that reduces mislabeling. We also leverage the integrated gradients feature attribution method~\cite{integrad_mukun17} to offer insights into which explanatory features most influenced each prospectivity prediction. We model prospectivity as a presence likelihood paired with an uncertainty. We consider epistemic uncertainty, measured based on the Monte-Carlo Dropout~\cite{mcdropout_gal16} method. In the following sections, we detail our DL MPM approach.

\begin{figure}[t]
  \centering
  \includegraphics[width=0.8\linewidth]{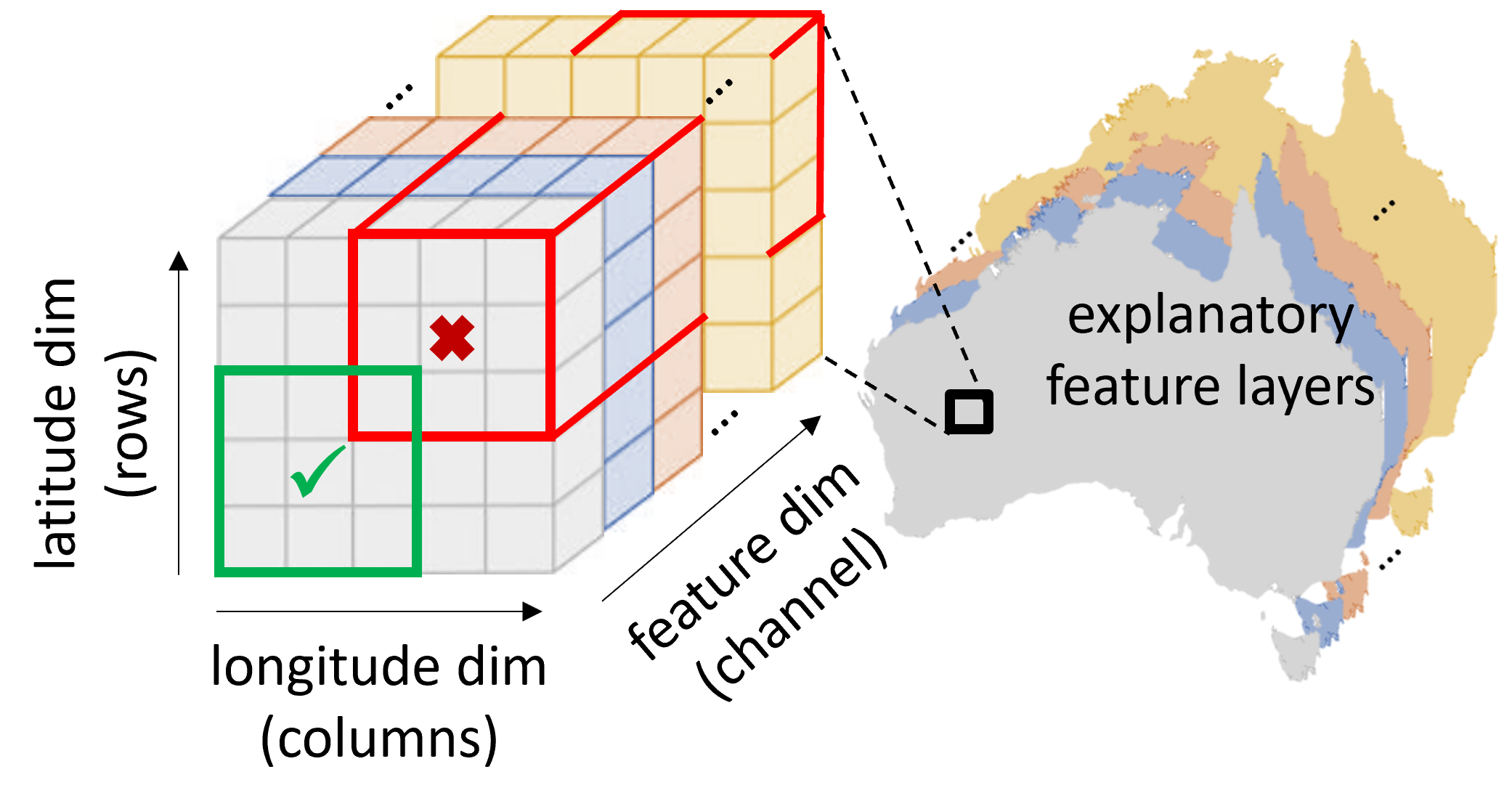}
  \vspace{-0.5cm}
  \caption{Illustration of data processing with multi-band georeferenced raster $X$. Red and green prisms are samples $x'_i \in \mathbb{R}^{m \times w \times w}$, with red \xmark and green \cmark representing absent and present center pixel labels for the samples, respectively.}
  \label{fig:preprocess}
  \Description{}
  \vspace{-0.5cm}
\end{figure}

\subsection{Data Preparation}\label{ss:preprocess}

We preprocess and sample our multi-band georeferenced raster $X$ to generate inputs for the entire training process of our approach (i.e., both supervised and self-supervised). Given a set of $m$ explanatory feature layers, we preprocess each using a series of operations common in ML MPM works~\cite{LAWLEY2022104635}. First, we rasterize and georeference each explanatory feature layer $j=\{1,\ldots,m\}$. Next, pixels $x_i$ that were outliers or had missing values were removed or imputed, respectively. Tukey fences were used to remove outliers, while inverse distance weighting on surrounding pixel values followed by smoothing were used for imputation. Last, the $m$ rasters are normalized to standard scores and geospatially aligned, for producing a multi-band georeferenced raster $X \in \mathbb{R}^{m \times r \times c}$. After the above preprocessing, the multi-band raster $X$ can be indexed with square windows of size $w$ to produce samples for training. A sample $x' \in \mathbb{R}^{m \times w \times w}$ from $X$ represents a 3D tensor, two dimensions for spatial extents (i.e., $w \times w$) and one for the explanatory feature layers $m$ as shown in Figure~\ref{fig:preprocess}.

\subsection{Self-Supervised Learning (SSL)}\label{ss:SSL}

We use masked image modeling within our multi-band georeferenced raster framework to pretrain a deep learning model, that serves as our backbone for feature extraction in downstream MPM tasks. Our approach is inspired by self-supervised learning via masked image modeling, which has had recent success within computer vision~\cite{he2022masked}. In masked image modeling, a visual deep neural network is pretrained by reconstructing input images that have severe signal loss (i.e., $\sim$75\% input image pixels discarded). In the context of MPM, our input images correspond to samples $x'$ of $X$, where the spatial extents of $x'$ are image widths and heights of size $w$ while the explanatory feature layers are image channels $m$. After pretraining, we use the backbone neural network for feature extraction in downstream MPM tasks.

\begin{figure}[t]
  \centering
  \includegraphics[width=0.98\linewidth]{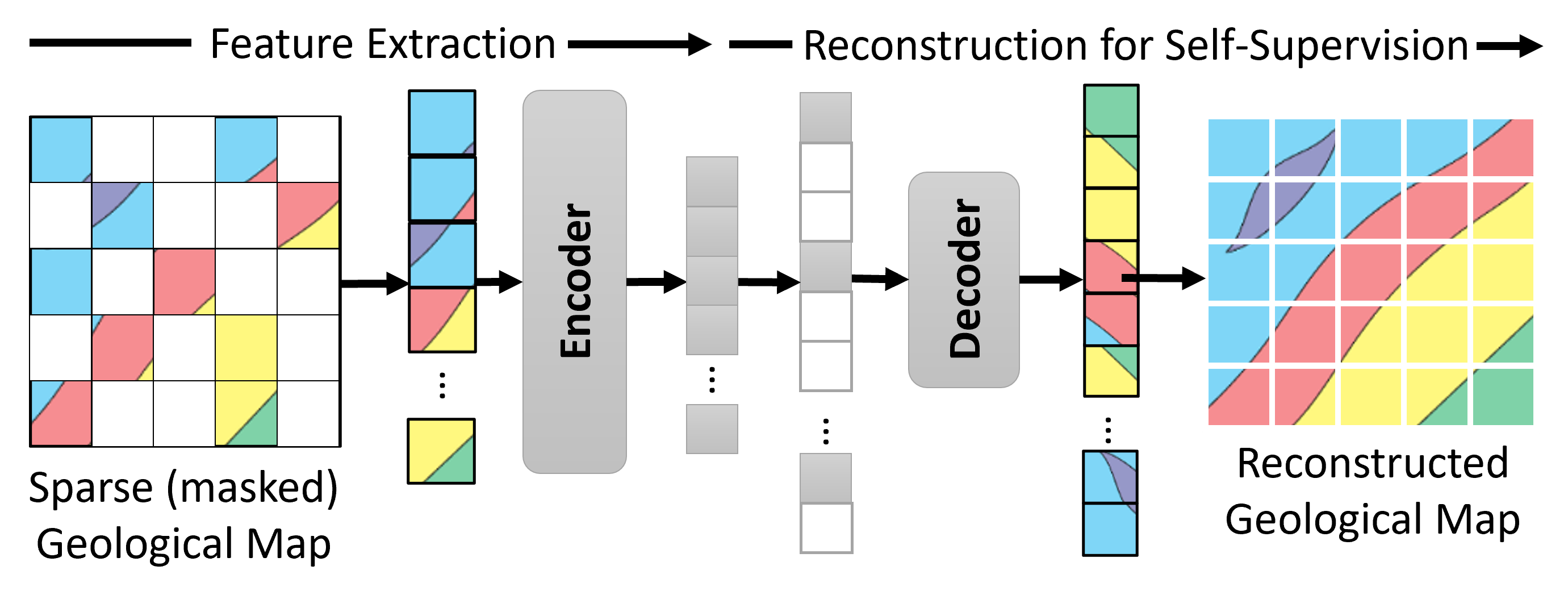}
  \vspace{-0.5cm}
  \caption{Illustration of masked image modeling pretraining procedure for a single explanatory feature layer.}
  \label{fig:ssl}
  \Description{}
  \vspace{-0.25cm}
\end{figure}

Our self-supervised model is an encoder-decoder architecture pretrained to reconstruct windows within a multi-band georeferenced raster $x'$ given their partial observations, see Figure~\ref{fig:ssl}. First the input sample is divided into regular non-overlapping patches. A small minority of the the patches are kept (i.e., $\sim$25\%) while the remainder are masked. The unmasked patches are sequenced and processed into unmasked patch tokens by an encoder $\mathcal{E}$, which in our architecture is a Vision Transformer~\cite{dosovitskiy2020vit}. Next, the unmasked patch tokens are supplemented with mask tokens that indicate which patches of the sample $x'$ were masked. These mask tokens are a learned representation indicating missing values that need to be predicted. The sequence of both unmasked patch tokens and mask tokens are then processed by the decoder, a second shorter series of transformer blocks. The output of the decoder is a sequence of patches that are stitched back together to reconstruct the input image. The Mean Squared Error (MSE) between reconstructed image from the decoder $\hat{x'}$ and the original input sample $x'$ is then the reconstruction objective used for pretraining, shown in Equation~\ref{eq:mse}. 

\begin{equation}\label{eq:mse}
    \mathcal{L}_{SS}(x_i) = \frac{1}{m \times w \times w} \sum_j (x'_i - \hat{x'_i})^2
\end{equation}

Due to the asymmetric encoder-decoder design of this masked image modeling technique~\cite{he2022masked}, the decoder has much few parameters and modeling capacity than the encoder. Therefore, the encoder $\mathcal{E}$ is optimized to learn a robust latent representation that summarizes the complete set of image patches when only a small portion of unmasked patches are observed. The learned features are a generic representation of the localized geospatial information that can be used for MPM. The encoder $\mathcal{E}$ alone can be used as the backbone network for feature extraction after pretraining.

\begin{figure}[t]
  \centering
  \includegraphics[width=0.98\linewidth]{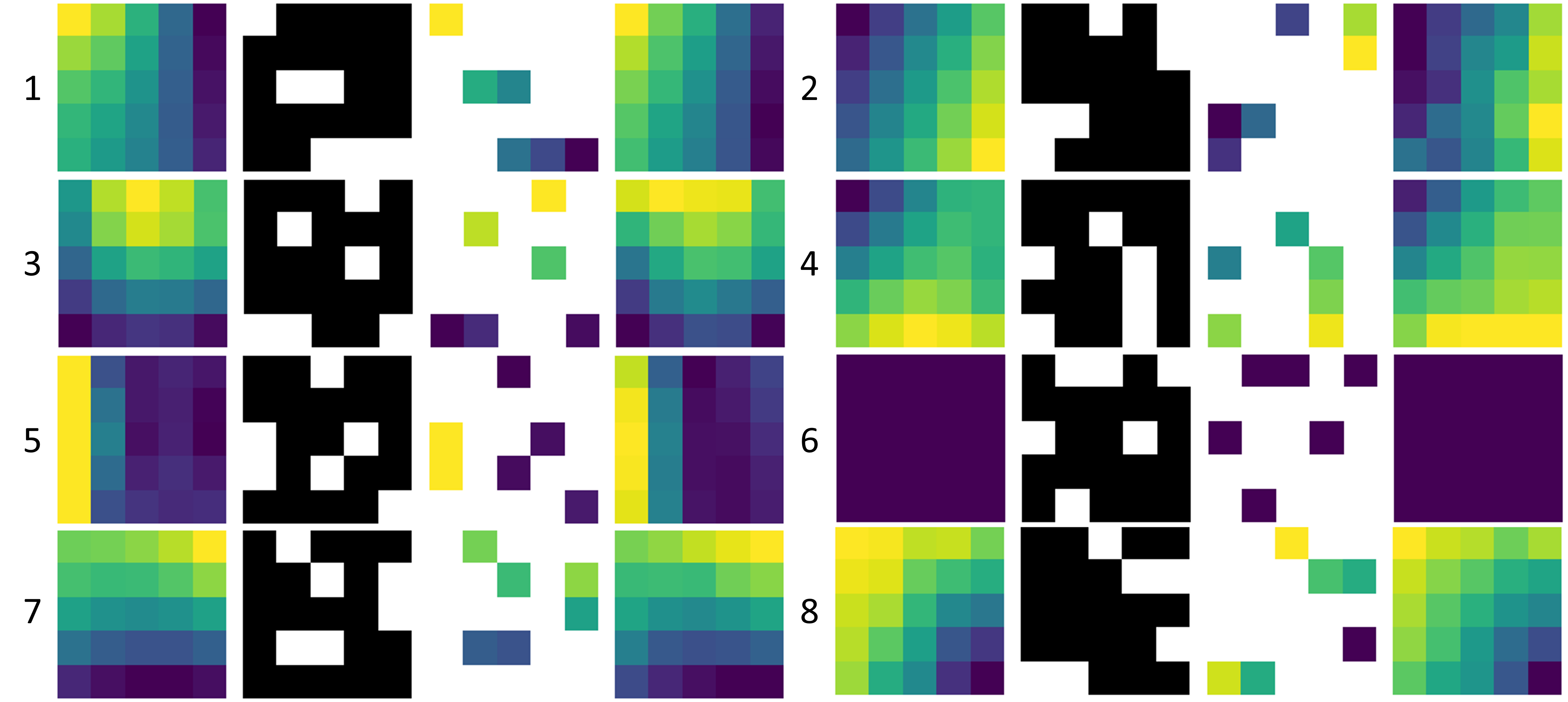}
  \vspace{-0.25cm}
  \caption{Eight masked image modeling pretraining examples showing input, 75\% mask, masked input, and reconstruction left-to-right in columns.}
  \label{fig:ssl_ex}
  \Description{}
  \vspace{-0.25cm}
\end{figure}

Figure~\ref{fig:ssl_ex} shows various inputs and outputs from the self-supervised pretraining as examples. In each of the eight examples the first column shows the original input, the second column shows the 75\% mask, the third column shows the sparse input after masking, and the fourth column shows the reconstructed input. Each example is taken from some explanatory feature layer that forms the samples $x'_i$. These explanatory feature layers include geophysical, geological, and geochemical raster data. While we visualize the examples as single bands in the figure, the masked image modeling occurs simultaneously across all $m$ explanatory feature layers. In this way, the learned features build spatial and channel redundancy.

\subsection{Mineral Prospectivity Mapping}\label{ss:mpm}

We consider MPM an imbalanced, positive-unlabeled binary classification problem. We formulate a new undersampling method for MPM by treating the samples that exclude those with exploration targets present as unlabeled data rather than negatives in Section~\ref{sss:pu_learn}. Our DL MPM approach combines the encoder $\mathcal{E}$ of the self-supervised network from Section~\ref{ss:SSL} with a small classifier to form the function $\mathcal{F}$ that estimates prospectivity. We discuss how we train $\mathcal{F}$ to output prospectivity in Section~\ref{sss:supervised_training}. Section~\ref{sss:xai} explains how we use feature attributions to compute which input features most influenced each prediction.

\subsubsection{Positive-Unlabeled Learning} \label{sss:pu_learn}

We treat MPM as an imbalanced, positive-unlabeled binary classification problem. MPM is known to be an imbalanced learning problem~\cite{prado2020modeling}. Prior MPM work assumes absence (i.e., negative samples $x'_{neg}$) for all remaining locations after excluding those with historical record of the exploration target being present (i.e., positive samples $x'_{pos}$). Typically, dataset balancing process then takes the form of some combination of random undersampling the majority class, $x'_{neg}$, and random or synthetic oversampling the minority class, $x'_{pos}$~\cite{zuo2020geodata}. However, this assumption (negative samples are the difference between the set of all samples and positive samples) increases the risks in labeling unknown true positives as negatives, i.e., $X\!\not\; x'_{pos} \neq x'_{neg}$. We instead introduce samples with unknown labels, $X\!\not\; x'_{pos} = x'_{unk}$, and propose a new undersampling approach to reduce this risk.

Our approach reduces the chance of labeling unknown true positives as negatives, by taking into account the feature similarities between positive $x'_{pos}$ and unknown $x'_{unk}$ samples. First, features for every sample in the MPM dataset are extracted. Sample features can be extracted directly from the $m$ explanatory feature layers of the multi-band georeferenced raster. We used the encoder $\mathcal{E}$ to perform feature extraction. Specifically, we use the generic latent representation outputted from the encoder to process each sample $x'$ from $X$. The features for all samples $x'$ are then split into positive and unknown sample features according to $X\!\not\; x'_{pos} = x'_{unk}$. Then we use a distance metric (e.g., Euclidean) to compute the similarity between each unknown sample $x'_{unk}$ and all available positive samples $x'_{pos}$. These distance computations result in a scale of unknown samples ordered by their similarity to the set of positive samples. Finally, we perform random undersampling for samples that fall within a range on the computed positive similarity scale, which is treated as a hyperparameter. Empirically, we have found that filtering 5-10\% of unknown samples most similar to the positive samples from undersampling to be effective.

\subsubsection{Estimating Prospectivity} \label{sss:supervised_training}

We treat MPM as a binary classification problem in which we seek to estimate the likelihood for the presence of some mineral exploration target at discrete locations. In our multi-band georeferenced raster framework, discrete locations correspond to individual pixels $x_i$ within the multi-band raster $X$. We pair the encoder $\mathcal{E}$ from Section~\ref{ss:SSL} with small classifiers to form the architecture $\mathcal{F}$ that predicts prospectivity. For practical use of our approach in mineral commodity assessments, we interpret prospectivity as providing both a presence likelihood and its uncertainty. In this work, we consider the epistemic uncertainty associated with the predictions of $\mathcal{F}$.

Discrete locations within our multi-band georeferenced raster framework are individual pixels $x_i \in \{x_1,\ldots,x_n\}$ within the multi-band raster $X$. Samples $x'_i$ for each pixel are formed from indexing square windows of size $w$ around $x_i$. These samples, $x'_i\in \mathbb{R}^{m \times w \times w}$, are 3D tensors that represent the local geospatial information, such as geophysical measurements, for $x_i$. Each $x_i$ is also accompanied by a label $y_i$ indicating whether the mineral exploration target is present, absent, or unknown. Note $y_i$ is a label for the center pixel within $x'$. The presence labels are determined by indexing the raster grid according to historical records of observed mineral deposits. In our approach, the absent and unknown labels are determined by undersampling method presented in Section~\ref{sss:pu_learn}.

For each exploration target, we pair the encoder $\mathcal{E}$ from Section~\ref{ss:SSL} with a small classifier to form the architecture that is trained with labeled samples. The combined architecture forms the function $\mathcal{F}$ from Section~\ref{ss:problem_defn} that estimates prospectivity. We train $\mathcal{F}$ on the subset of samples labeled either present or absent; unknown samples are used to make continuous prospectivity maps and excluded in training. Our classifier consists of a Multi-Layer Perceptron (MLP) with parametric rectified linear units for activation layers and BatchNorms for normalization layers. As input, the classifier processes the single-dimensional latent features extract by the encoder $\mathcal{E}$. The classifier outputs a single scalar $\hat{y}_i$ estimating the presence likelihood at $x_i$, $\hat{y}_i = \mathcal{F}(x'_i) \approx p(y_i=\textrm{present} | x'_i)$. We use the Binary Cross-Entropy (BCE) between the predictions $\hat{y}_i$ and labels $y_i$ as the supervised training objective $\mathcal{L}_{S}$, shown in Equation ~\ref{eq:bce} where $j$ indexes only samples labeled as present or absent and $J$ is the number of such samples.
\begin{equation}\label{eq:bce}
    \mathcal{L}_S\big(\hat{y}_i, y_i\big) = - \frac{1}{J} \sum_j y_i \cdot \log (\hat{y}_i) + (1-y_i) \cdot \log (1-\hat{y}_i)
\end{equation}

\begin{figure}[t]
  \centering
  \includegraphics[width=0.95\linewidth]{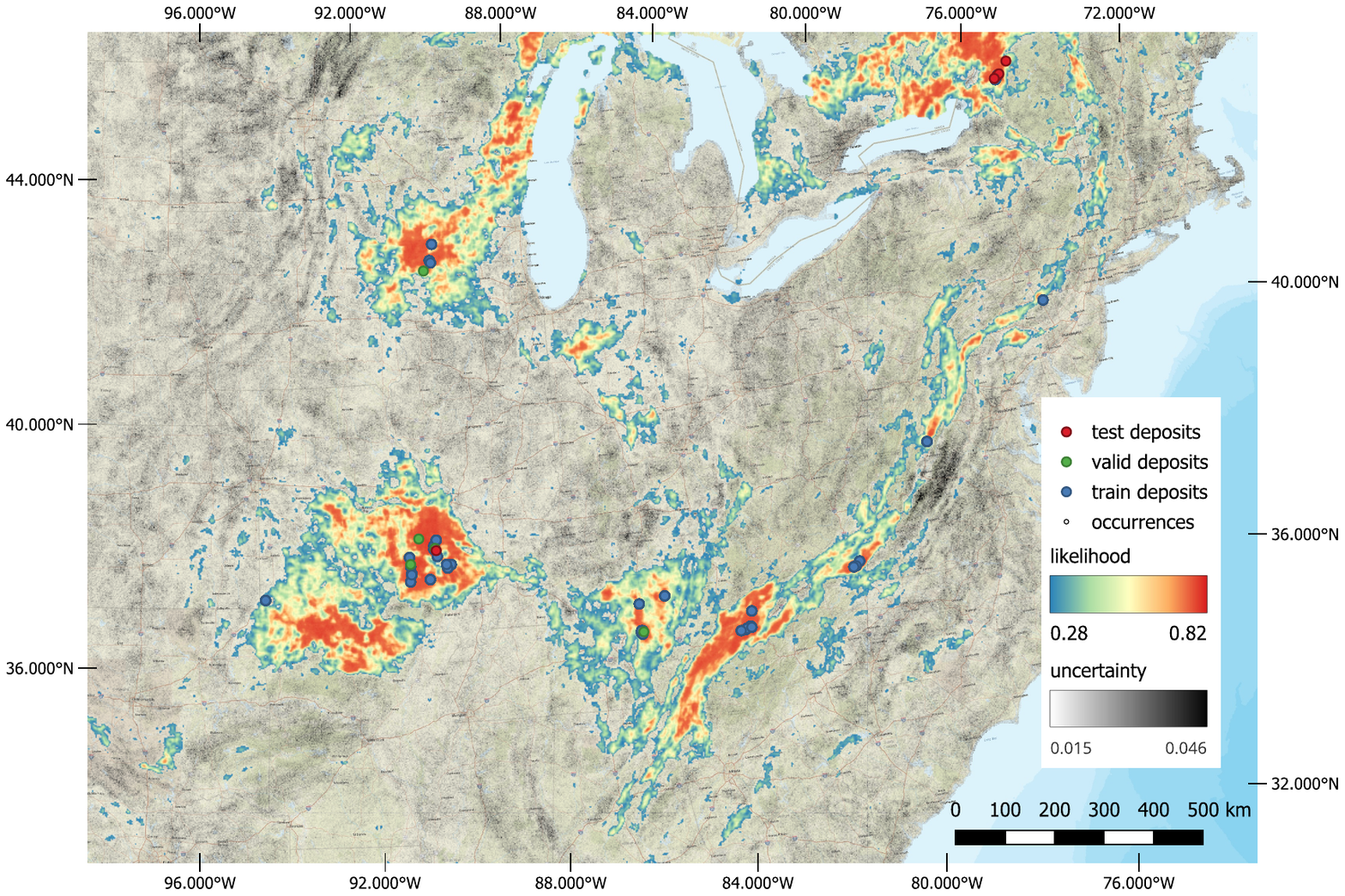}
  \vspace{-0.25cm}
  \caption{MVT Lead-Zinc prospectivity map showing likelihoods overlayed onto uncertainties. Deposits are grouped by usage: training, validation, or testing.}
  \label{fig:approach_mpm}
  \Description{}
  \vspace{-0.25cm}
\end{figure}
\begin{figure}[t]
  \centering
  \includegraphics[width=0.95\linewidth]{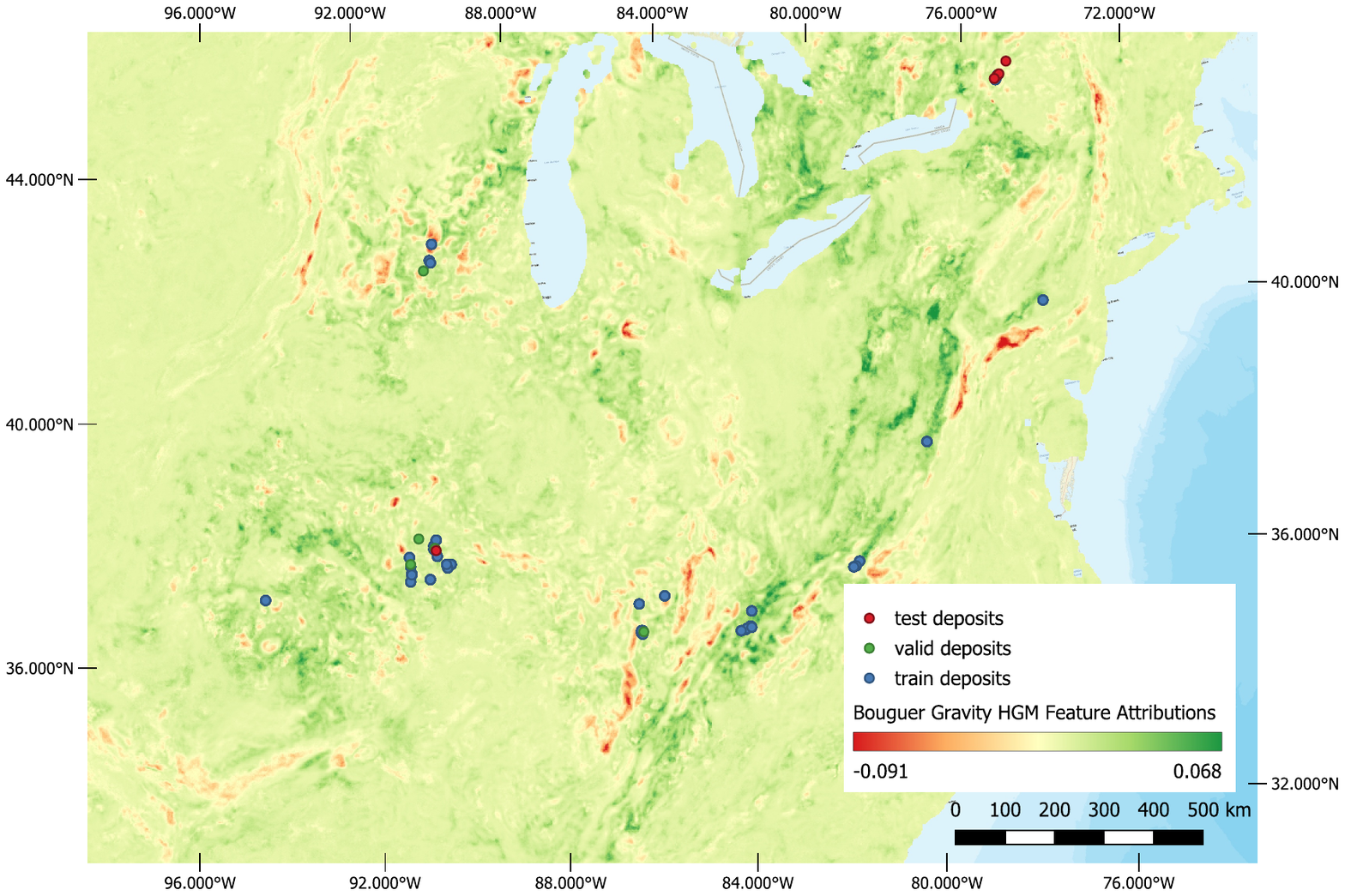}
  \vspace{-0.25cm}
  \caption{An example MVT Lead-Zinc feature importance map for a geophysical explanatory feature layer.}
  \label{fig:approach_IG}
  \Description{}
  \vspace{-0.5cm}
\end{figure}

Our method provides model uncertainty associated with each prediction using Monte-Carlo (MC) Dropout~\cite{mcdropout_gal16}. Epistemic uncertainty is often overlooked in prior ML MPM implementations, although uncertainties are inherently present within the MPM problem~\cite{zuo2021uncertainties}. Epistemic uncertainty is the reducible portion of a model's total predictive uncertainty, due to a lack of sufficient knowledge~\cite{hullermeier2021aleatoric}. Many sources of epistemic uncertainty exist, including insufficient data, low diversity of training examples, and others~\cite{gawlikowski2023survey}. We use MC Dropout to provide the epistemic uncertainty within the MPM predictions associated with the learned parameters of $\mathcal{F}$~\cite{mcdropout_gal16}. To implement MC Dropout, our MLP classifier contains Dropout within each layer. The Dropouts are kept active both at training and inference time to estimate the mean and variance of the predictive distribution, the presence likelihood. The mean and variance of the presence likelihood are estimated using the sample mean and variance over $T$ stochastic forward passes of the classifier.

Using the above formulation, our DL MPM approach outputs two georeferenced rasters which represent the prospectivity mean and standard deviation, shown in Figure~\ref{fig:approach_mpm}. We use our trained classifier to compute the mean and variance of the exploration target presence likelihood for every sample $x'_i$. By reverse-indexing the samples $x'_i$ to their corresponding pixels $x_i$ within the multi-band georeferenced raster $X$, continuous maps of the presence likelihood mean and standard deviation can be generated. Figure~\ref{fig:approach_mpm} shows the eastern United States MVT Lead-Zinc prospectivity maps generated in our experiments. The mean prospectivity is visualized as a heatmap (red color as highest value), while the standard deviation is in gray-scale (black color as highest value).

\subsubsection{Explaining DL MPM Predictions}\label{sss:xai}

It is important for domain experts to interpret and verify the predictions from DL methods. However, validation of DL generated prospectivity maps from a geological perspective is challenging due to the complex nature of these techniques.  Prior ML MPM work, such as WoE~\cite{agterberg1989systematic} and RF~\cite{rodriguez2014predictive}, are limited to providing aggregate explanatory variable importance over many predictions. Variable importance on every input feature for individual predictions can better help validate individual predictions spatially, in addition to relative importances between explanatory variables in the aggregate. As the domain knowledge about mineralization continues to evolve, being able to assess explanatory variable importance spatially could provide new insights from which correlations get highlighted.

We use an explainable artificial intelligence technique, Integrated Gradients (IG) \cite{integrad_mukun17}, to compute importance scores for every input feature to a DL MPM prediction. IG is a method used to compute which input features contribute most to the prediction made by the DL model, thus improving the interpretability. For an input sample $x'$, we compute the gradient of the network with respect to $x'$ along a path from a baseline sample to $x'$. The integral of these gradients provides the importance score for each feature. The baseline sample has the same dimensionality as $x'$ but serves as a reference devoid of any input signal (i.e., a zero-tensor). Similar to prospectivity, we can reverse-index the samples $x'_i$ to their corresponding pixels $x_i$ within the multi-band georeferenced raster $X$, for generating continuous maps of feature importance for all $m$ explanatory feature layers. In Figure~\ref{fig:approach_IG}, we provide the feature importance map for an explanatory variable as a heatmap (green color as the high value). The signs and magnitudes of IG importance scores indicate the type of association and strength of contribution, respectively, the input has on the prediction. Positive IG values increase the presence likelihood while negative values decrease it.

\section{Evaluation}\label{s:evaluation}

We validated our MPM method and performed additional ablation studies to substantiate our design choices. For our experiments, we followed the procedures in~\cite{LAWLEY2022104635} to recreate datasets for mineral assessments of Mississippi Valley-Type (MVT) and Clastic-Dominated (CD) Lead-Zinc deposits in North America and Australia. We compared the prospectivity mapping performance of our approach with five baseline methods: weights of evidence~\cite{agterberg1989systematic}, gradient boosting~\cite{LAWLEY2022104635}, artificial (i.e., feed-forward) neural networks~\cite{brown2000artificial}, convolutional neural networks~\cite{sun2020data}, and transformers (self-attention)~\cite{yin2023mineral}. A suite of classification metrics were tracked to confirm consistency in results. Results in each experiment were reported as the average and standard deviation of 5 repeated trials with different training splits. To better contextualize the prediction results, we also reported an analysis of each model's computational complexity. Ablation studies were provided to verify individual components, including self-supervision and undersampling, from our approach.

\begin{table}[t]
    \caption{Explanatory Feature Layers from \cite{LAWLEY2022104635}}
    \label{tab:evidence}
    \vspace{-0.25cm}
    \begin{tabular}{l|l}
        \toprule
        Lithology (majority)                       & Proximity to black shales                            \\
        \hline
        Lithology (minority)                       & Proximity to terrane boundaries                      \\
        \hline
        Period (maximum)                           & Proximity to passive margins                         \\
        \hline
        Period (minimum)                           & Sedimentary dictionaries                             \\
        \hline
        Gravity Bouguer                            & Metamorphic dictionaries                             \\
        \hline
        Igneous dictionaries                       & Gravity upward-continued HGM                         \\
        \hline
        Gravity HGM                                & Gravity worms                                        \\
        \hline
        Depth to LAB                               & Gravity upward-continued worms                       \\
        \hline
        Depth to Moho                              & Magnetic HGM                                         \\
        \hline
        Paleo-latitude                             & Magnetic long-wavelength HGM                         \\
        \hline
        Proximity to faults                        & Magnetic worms                                       \\
        \hline
        Satellite gravity                          & Magnetic long-wavelength worms                       \\
        \bottomrule
    \end{tabular}
    \vspace{-0.5cm}
\end{table}

\subsection{MPM Experimental Setup}\label{ss:experimental_setup}

We formed our MPM datasets using the explanatory feature layers and deposit records compiled and released in~\cite{LAWLEY2022104635}. The authors selected a set of explanatory feature layers and deposit records to perform mineral assessments of MVT and CD Lead-Zinc deposits in North America and Australia. The comprehensive set of explanatory features included a variety of modalities spanning geophysics (e.g., magnetic, gravity, and seismic anomaly), geology (e.g., age, lithology), and geochemistry (e.g., black shale proximity), detailed in Table~\ref{tab:evidence}. We used the same set of explanatory features, because selecting a different (better) set of features is beyond the scope of this work. Our experiments differ from~\cite{LAWLEY2022104635} in that we emphasize comparisons of the ML MPM algorithms with a comprehensive set of baselines and metrics. As what we described in Section~\ref{ss:preprocess}, we compiled the $m$ explanatory feature layers into a multi-band georeferenced raster $X \in \mathbb{R}^{m \times r \times c}$ with $r$ rows and $c$ columns. 

Similar to $X$, we formed single-band georeferenced rasters $Y \in \mathbb{R}^{1 \times r \times c}$ for each mineral deposit type. The default label for pixels in $Y$ is unknown. Each label raster $Y$ is then updated to write the present labels by georeferencing historical records of the deposit type to index the pixels within the raster grid. After rasterizing historical records, our dataset across North America and Australia contains 174 MVT and 77 CD mineral deposit sites (pixels). Each label raster $Y$ is then updated again to write the absent labels using our undersampling approach presented in Section~\ref{sss:pu_learn}. The present and absent pixels in $Y$ are used to train the MPM approach for each deposit type. In addition to our undersampling approach, we used oversampling to ensure the binary labels in training sets were balanced in all experiments. The present and absent label pixels $y_i$ in $Y$ are aligned with explanatory feature pixels $x_i$ in $X$ (i.e., georeferencing in $X$ corresponds to $Y$). Hence, the present and absent label pixels $y_i$ can be used to index the corresponding explanatory feature pixels $x_i$. Square windows of size $w$ surrounding $x_i$ serve as our input samples $x'_i$ accompanying labels.

\begin{table*}
  \caption{MVT Results; for all the metrics the higher results, the better; mean $\pm$ std over 5 random seeds}
  \label{tab:mvt_results}
  \vspace{-0.25cm}
  \begin{tabular}{l|cccccc}
    \toprule
    \bf{Baseline} & \bf{F1} & \bf{MCC} & \bf{AUPRC} & \bf{B.ACC} & \bf{AUROC}$^\dagger$ & \bf{ACC}$^\dagger$ \\
    \midrule
    WoE & 51.4$\pm$10.3 & 51.0$\pm$9.2 & 60.8$\pm$7.5 & 76.8$\pm$8.9 & 95.8$\pm$1.2 & 94.9$\pm$1.5 \\
    \hline
    GBM & 76.0$\pm$3.1 & 75.2$\pm$3.3 & 83.8$\pm$4.4 & 85.7$\pm$2.9 & \textbf{97.7$\pm$1.7} & 97.8$\pm$0.3 \\
    \hline
    ANN & 74.7$\pm$5.9 & 73.9$\pm$5.9 & 83.7$\pm$3.9 & 85.6$\pm$5.7 & 97.6$\pm$1.5 & 97.7$\pm$0.4 \\
    \hline
    CNN & 71.4$\pm$6.9 & 70.4$\pm$7.4 & 78.4$\pm$12.2 & 83.3$\pm$2.2 & 93.8$\pm$2.5 & 97.3$\pm$0.9 \\
    \hline
    ViT & 54.9$\pm$11.9 & 53.1$\pm$12.2 & 64.8$\pm$6.5 & 75.2$\pm$7.4 & 95.2$\pm$2.6 & 95.9$\pm$0.9 \\
    \hline
    Ours & \textbf{80.3$\pm$6.6} & \textbf{79.7$\pm$7.0} & \textbf{84.3$\pm$7.6} & \textbf{89.0$\pm$2.0} & 96.8$\pm$2.7 & \textbf{98.0$\pm$0.8} \\
    \bottomrule
    \multicolumn{7}{r}{\footnotesize{$\dagger$ Metrics not intended for imbalanced datasets.}} \\ 
  \end{tabular}
  \vspace{-0.25cm}
\end{table*}
\begin{table*}
  \caption{CD Results; for all the metrics the higher results, the better; mean $\pm$ std over 5 random seeds}
  \label{tab:cd_results}
  \vspace{-0.25cm}
  \begin{tabular}{l|cccccc}
    \toprule
    \bf{Baseline} & \bf{F1} & \bf{MCC} & \bf{AUPRC} & \bf{B.ACC} & \bf{AUROC}$^\dagger$ & \bf{ACC}$^\dagger$ \\
    \midrule
    WoE & 45.1$\pm$15.2 & 46.7$\pm$15.0 & 56.4$\pm$11.7 & 68.2$\pm$8.5  & 90.2$\pm$6.3 & 95.8$\pm$0.9 \\
    \hline
    GBM & 52.1$\pm$14.7 & 52.8$\pm$14.0 & 73.6$\pm$11.0 & 80.4$\pm$12.8 & 97.5$\pm$1.2 & 94.3$\pm$2.7 \\
    \hline
    ANN & 64.8$\pm$13.9 & 63.7$\pm$14.3 & 81.9$\pm$10.8 & 81.5$\pm$7.2  & 98.3$\pm$1.2 & \textbf{96.5$\pm$1.8} \\
    \hline
    CNN & 70.7$\pm$7.2  & 70.9$\pm$6.5  & 84.7$\pm$4.9  & 93.2$\pm$2.2  & 98.4$\pm$1.3 & 96.2$\pm$1.3 \\
    \hline
    ViT & 62.4$\pm$11.8 & 62.5$\pm$12.0 & 64.9$\pm$14.8 & 80.2$\pm$6.5  & 91.3$\pm$7.5 & 96.2$\pm$1.5 \\
    \hline
    Ours & \textbf{75.3$\pm$16.4} & \textbf{76.3$\pm$15.4} & \textbf{93.7$\pm$5.1}  & \textbf{93.4$\pm$7.3}  & \textbf{99.4$\pm$0.6} & 96.4$\pm$3.5 \\
    \bottomrule
    \multicolumn{7}{r}{\footnotesize{$\dagger$ Metrics not intended for imbalanced datasets.}} \\ 
  \end{tabular}
  \vspace{-0.25cm}
\end{table*}

After forming the present and absent labels $y_i$ and their corresponding samples $x'_i$, we generate our ML MPM datasets to train all methods, including ours and five baseline methods. From the complete set of labels $y_i$ and samples $x'_i$, we split the data using the 80/10/10 rule to form the mutually exclusive train, validation, and test sets of our data. We repeated this splitting with different random seeds and reported results as the average and standard deviation across the random seed trials.

We measured a suite of metrics to evaluate the prediction performance of each ML MPM method, including \textbf{F1-score}, \textbf{A}rea \textbf{U}nder the \textbf{R}eceiver \textbf{O}perating \textbf{C}haracteristic Curve (\textbf{AUROC}), \textbf{A}rea \textbf{U}nder the \textbf{P}recision-\textbf{R}ecall \textbf{C}urve (\textbf{AUPRC}), \textbf{Acc}uracy (\textbf{ACC}), \textbf{B}alanced \textbf{Acc}uracy (\textbf{B.ACC}), and \textbf{M}atthews \textbf{C}orrelation \textbf{C}oefficient (\textbf{MCC}). F1-score provides a balanced measure of a model's precision and recall, making it particularly useful for evaluating performance on imbalanced datasets where both false positives and false negatives matter. The AUROC measures the ability of the model to distinguish between classes by plotting recall against the false positive rate, indicating how well the model can distinguish between positive and negative classes. The AUPRC evaluates the trade-off between precision and recall for different classification thresholds. The area under the precision-recall curve reflects the model's performance across all recall levels, being particularly useful for imbalanced datasets. ACC measures the overall correctness of the model by calculating the proportion of true results (both true positives and true negatives) among the total number of outcomes (i.e., true and false, positives and negatives). B.ACC is defined as an average of the true positive and true negative rates. It accounts for imbalanced class distributions and provides a more accurate measure of a model's performance across both classes. The MCC is a comprehensive metric that considers all possible outcomes and provides a correlation coefficient between actual and predicted binary classifications, especially effective with imbalanced data.

\begin{figure}[t]
  \centering
  \includegraphics[width=0.98\linewidth]{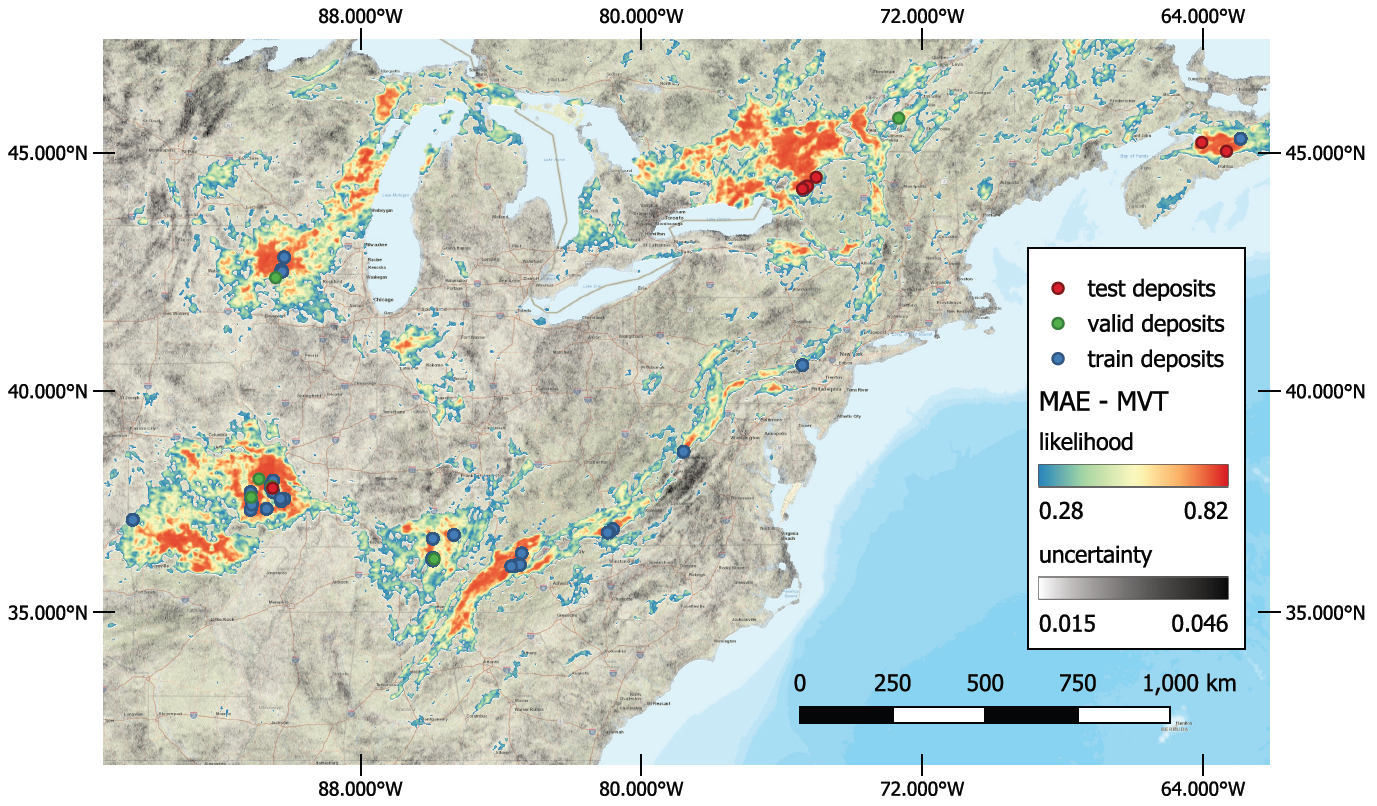}
  \vspace{-0.25cm}
  \caption{Prospectivity from MAE showing gradual likelihood gradients; uncertainties are non-uniform.}
  \label{fig:mvt_midcont_mae}
  \Description{Description goes here; less than 2000 characters long (including spaces)}
  \vspace{-0.5cm}
\end{figure}

We compared the prospectivity mapping performance of our approach with five prior MPM methods. The Weights of Evidence (\textbf{WoE}) baseline calculates the log-odds ratios based on observed associations between predictor variables and the target outcome~\cite{agterberg1989systematic}. These log-odds ratios are then integrated to assess the likelihood of the deposit. The Gradient Boosting Machine (\textbf{GBM}) baseline builds a strong predictive model by sequentially adding weaker models, decision trees, each of which corrects the errors of its predecessors~\cite{LAWLEY2022104635}. The feed-forward (Artificial) Neural Networks (\textbf{ANN}) baseline was implemented as an MLP architecture identical to our classifier MLP in Section~\ref{sss:supervised_training}~\cite{brown2000artificial}. The ANN is trained by optimizing a loss function through gradient descent. The Convolutional Neural Networks (\textbf{CNN}) baseline builds on the ANN by incorporating convolutional layers that automatically and adaptively learn spatial hierarchies of features~\cite{sun2020data}. We used ResNet10t~\cite{he2016deep} as our CNN because it is a deep network architecture with similar capacity to the other deep learning methods. The Transformers (\textbf{ViT}) baseline was implemented as a vision transformer architecture~\cite{dosovitskiy2020vit} using six ViT transformer blocks, each of which uses self-attention to process inputs~\cite{yin2023mineral}. The vision transformer works similar to a text transformer by treating an image as a sequence of smaller patches. The encoder architecture of our approach (\textbf{Ours}) was implemented identically to the ViT baseline and paired with an MLP classifier implemented with an architecture identical to the ANN baseline and detailed in Section~\ref{sss:supervised_training}.

\begin{figure}[t]
  \centering
  \includegraphics[width=0.98\linewidth]{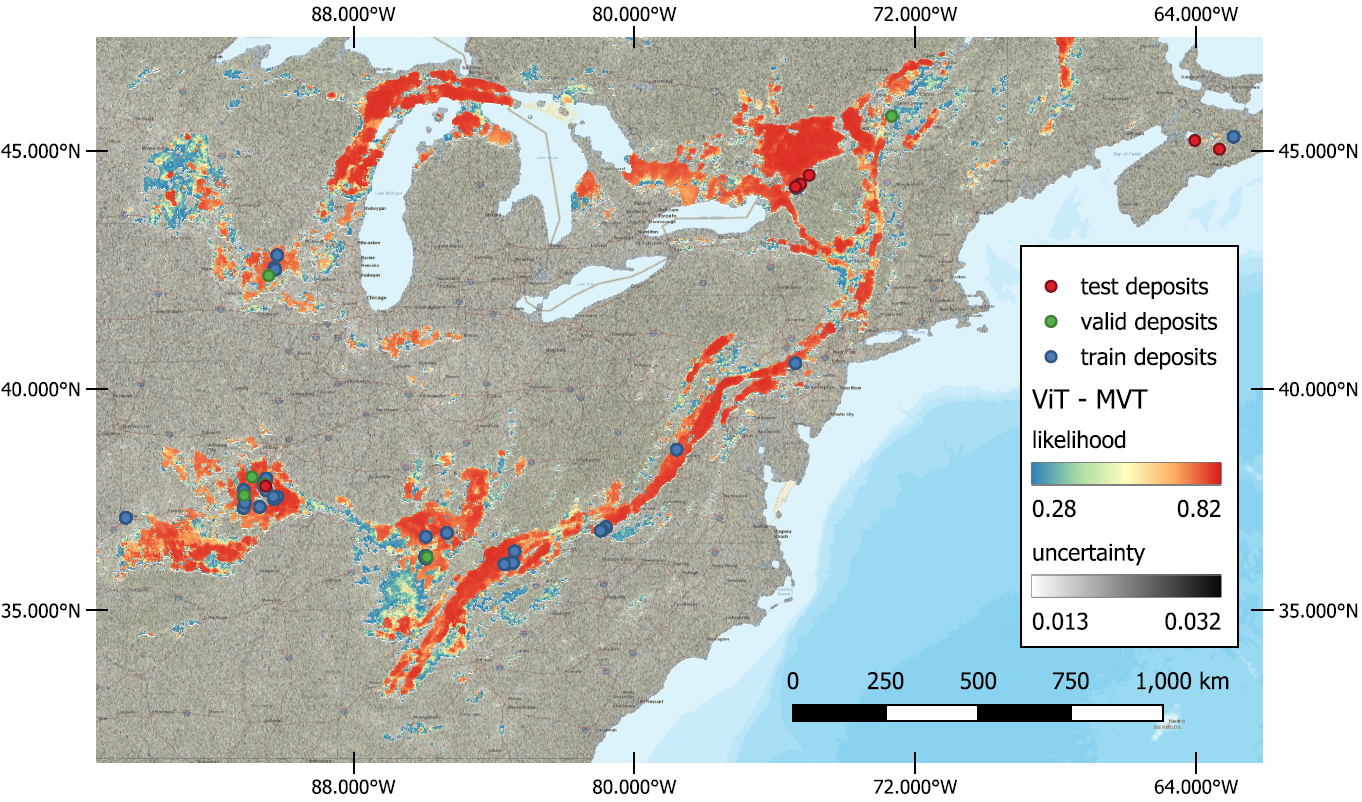}
  \vspace{-0.25cm}
  \caption{Prospectivity map from ViT showing overconfident predictions with uniformly high uncertainty across the map.}
  \label{fig:mvt_midcont_vit}
  \Description{Description goes here; less than 2000 characters long (including spaces)}
  \vspace{-0.5cm}
\end{figure}

\subsubsection{Pretraining}\label{sss:exp_pretrain}

Prior to beginning the mineral assessment experiments, we pretrained our DL MPM approach using self-supervised learning as described in Section~\ref{ss:SSL}. The pretraining data consisted of the all the samples $x'_i$ from the pixels $x_i$ (i.e., pixels labeled present, absent, and \textit{unknown}). Note that because the pretraining is self-supervised, no labels whatsoever are used during pretraining. Instead, the samples $x'_i$ serve as the input data and labels. The decoder used during pretraining consisted of two transformer blocks. During pretraining, we track two reconstruction metrics, Structural Similarity Index Measure (SSIM) and Peak Signal-to-Noise Ratio (PSNR), to determine when the network has been sufficiently trained. SSIM is a perceptual metric used to evaluate the visual similarity between two images by comparing their luminance, contrast, and structural information. PSNR is a metric used to measure the quality of a reconstructed or compressed image, by comparing the maximum possible signal value to the level of background noise. Empirically, we observed the pretraining reconstruction performance converged within 30 epochs. In addition to the reconstruction metrics, we qualitatively inspect a subset of the reconstructed images to confirm their fidelity to the original images. Examples are shown in Figure~\ref{fig:ssl_ex}. After pretraining, we use the encoder $\mathcal{E}$ from the encoder-decoder architecture for feature extraction. The encoder $\mathcal{E}$ only needs to be pretrained once because the explanatory feature layers do not change across the target deposit types.

\subsection{MPM Experimental Results}\label{ss:mpm_results}

The quantitative results of our MVT and CD Lead-Zinc deposit assessments are summarized by Tables~\ref{tab:mvt_results} and \ref{tab:cd_results}, respectively. Our approach consistently improved over the baselines across both mineral deposit types for the F1, MCC, AUPRC, and B.ACC prediction metrics. It is worth noting that F1, MCC, AUPRC, and B.ACC metrics are all designed for use with imbalanced datasets while AUROC and ACC are not. In an imbalanced testing dataset, AUROC and ACC can be less informative because the predictive performance on the absent samples overpowers the few present samples. Interestingly, ViT has lower prediction performance compared with Ours despite having identical network architectures. We believe the lack of pretraining causes the ViT baseline to overfit.

\begin{figure}[t]
  \centering
  \includegraphics[width=0.98\linewidth]{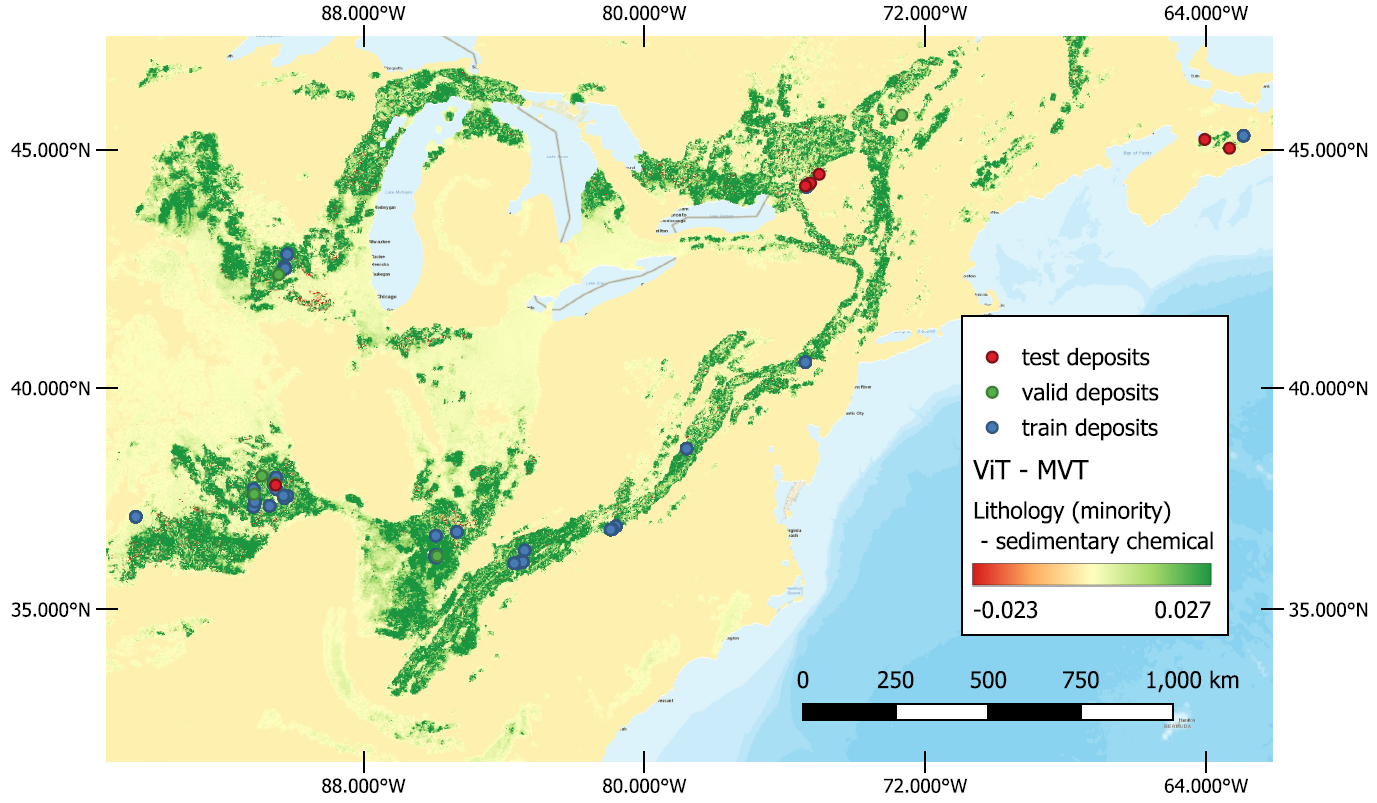}
  \vspace{-0.25cm}
  \caption{Feature importance map showing the ViT baseline is overfitting to surface lithology.}
  \label{fig:mvt_midcont_vit_attr}
  \Description{Description goes here; less than 2000 characters long (including spaces)}
  \vspace{-0.5cm}
\end{figure}

\begin{figure}[t]
  \centering
  \includegraphics[width=0.98\linewidth]{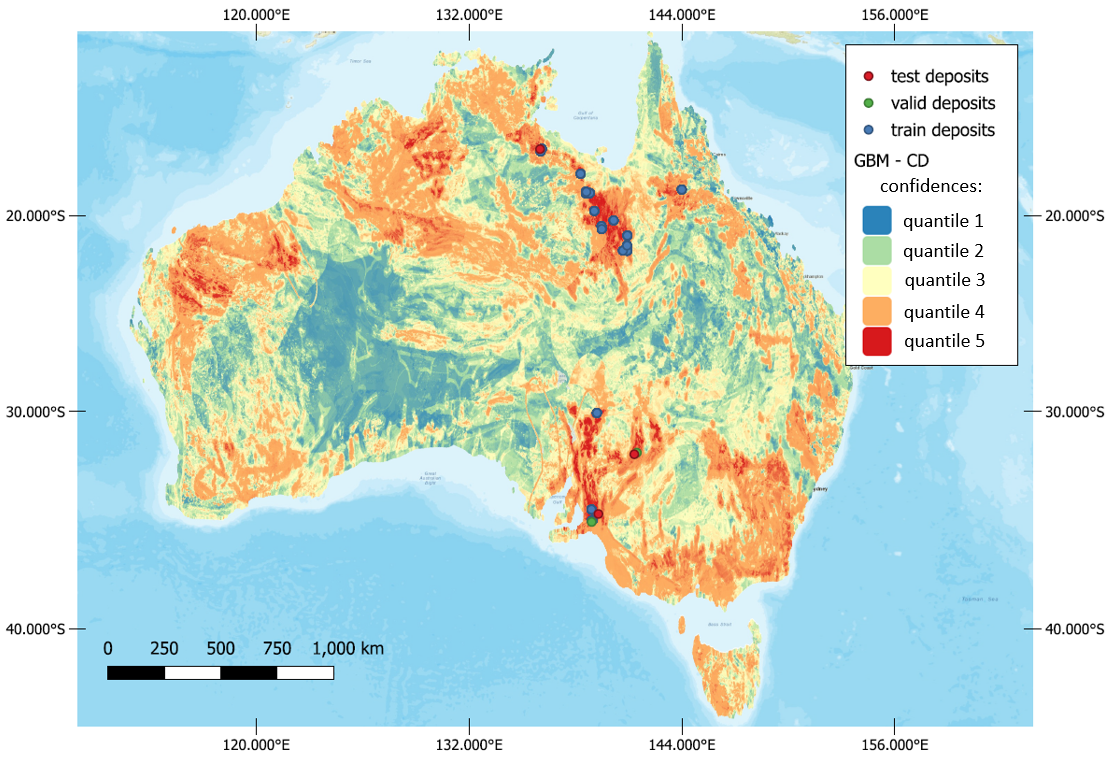}
  \vspace{-0.25cm}
  \caption{Prospectivity map from GBM visualizing five quantiles of model confidences as discrete colors. The 5th quantile is most perspective.}
  \label{fig:cd_aus_gbm}
  \Description{Description goes here; less than 2000 characters long (including spaces)}
  \vspace{-0.25cm}
\end{figure}

\begin{figure}[t]
  \centering
  \includegraphics[width=0.98\linewidth]{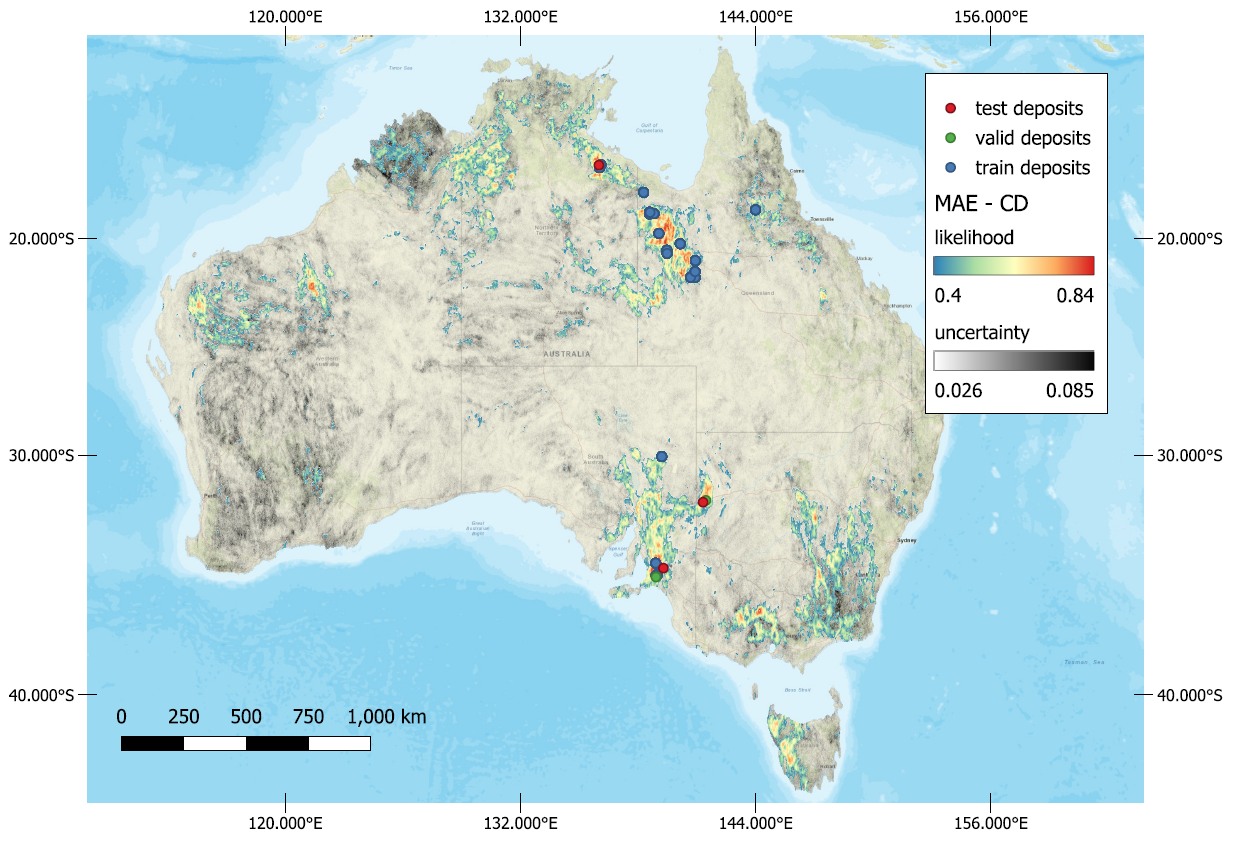}
  \vspace{-0.25cm}
  \caption{Prospectivity from MAE that shows agreement with most prospective areas in the GBM model, quantile 5, and uncertainties have similar structure to quantile 4.}
  \label{fig:cd_aus_mae}
  \Description{Description goes here; less than 2000 characters long (including spaces)}
  \vspace{-0.5cm}
\end{figure}

Some qualitative comparisons of the MVT and CD Lead-Zinc deposit assessments are shown in Figures~\ref{fig:mvt_midcont_mae} through~\ref{fig:cd_aus_mae}. In Figure~\ref{fig:mvt_midcont_vit}, we provide the prospectivity maps produced by the ViT baseline in NE United States and SE Canada for MVT Lead-Zinc deposits. Likelihoods produced by ViT tend to be overconfident and uncertainties are uniformly high. We also see that ViT did not predict several test and train deposits, including Missouri and Nova Scotia. We further analyze ViT, by considering the feature importance for its predictions, for which a single explanatory feature layer has been provided in Figure~\ref{fig:mvt_midcont_vit_attr}. From Figure~\ref{fig:mvt_midcont_vit_attr}, it is evident ViT is overfitting to the surface geology feature, Lithology (minority). Visually, the prospectivity maps from our approach, in Figure~\ref{fig:mvt_midcont_mae}, are more precise. Our predictions also include the deposits that ViT missed. Additionally, changes in presence likelihoods in our output are more gradual and uncertainties have become non-uniform with some areas having higher uncertainty, such as north of Lynchburg, VA. As another example for comparison, we consider the prospectivity maps produced by the GBM baseline and our approach for CD deposits in Australia shown in Figures~\ref{fig:cd_aus_gbm} and~\ref{fig:cd_aus_mae}, respectively. Note, the GBM baseline only outputs a confidence value for the prediction (no uncertainty). Therefore, we visualized the GBM prospectivity map using five quantiles to make better comparisons with the likelihoods and uncertainties in our prospectivity maps. From the visualization, the MAE and GBM often overlap on most prospective areas. Additionally, the uncertainty map of our approach shares a similar structure to the lower confidence quantiles of the GBM, such as the SW and SE corners of Australia.

As a final dimension for comparison, we consider the computational complexity in terms of the number of trainable parameters (params) and floating point operations (FLOPS) for each method. The WoE, GBM, and ANN baselines have negligible computational cost compared with the deep learning approaches. Considering the deep learning approaches, the CNN carries a slightly higher computational burden than the transformer models although CNN prediction performance tended to be worse in the experiments. The CNN has 4.94 million params and 5.50 million FLOPS for each prediction while the transformer models (i.e., ViT and Ours) have 4.80 million params and 5.28 million FLOPS. However, for our method, we can divide the computational burden between the backbone feature extraction network and classifier. Our method is only required to train the classifier during supervised training process, as the backbone features for all samples have been extracted during the pretraining stage. The feature extraction backbone of our method contributes 4.75 million params and 5.23 million FLOPS per prediction, while the classifier accounts for only 0.04 million params and 0.04 million FLOPS. Therefore, by using frozen features from the backbone feature extraction, we drastically improve the computational efficiency of training a specific classification task without sacrificing prediction performance.

\begin{table}[t]
    \caption{Summary of Feature Robustness Results; for all the metrics the higher results, the better; mean $\pm$ std over 5 random seeds}
    \label{tab:robust_abs}
    \vspace{-0.25cm}
    \begin{tabular}{l|c|c|c|c}
        \toprule
        \textbf{Deposit:} & \multicolumn{2}{c|}{\bf{MVT}} & \multicolumn{2}{c}{\bf{CD}} \\
        \hline
        \textbf{Baseline}   & \textbf{F1}           & \textbf{AUPRC}        & \textbf{F1}            & \textbf{AUPRC} \\
        \midrule
        WoE                 & 6.5$\pm$5.5           & 8.0$\pm$1.5           & 3.6$\pm$6.2            & 9.7$\pm$3.5 \\
        \hline
        GBM                 & 12.4$\pm$12.5         & 44.7$\pm$10.5         & 15.3$\pm$20.2          & 32.5$\pm$16.3 \\
        \hline
        ANN                 & 35.3$\pm$7.2          & 37.0$\pm$4.5          & 40.7$\pm$6.4           & 45.9$\pm$6.2 \\
        \hline
        CNN                 & 47.9$\pm$6.2          & 44.5$\pm$8.7          & 36.2$\pm$12.2          & 38.1$\pm$10.6 \\
        \hline
        ViT                 & 54.9$\pm$11.9         & 64.8$\pm$6.5          & 62.4$\pm$11.8          & 64.9$\pm$15.9 \\
        \hline
        Ours                & \textbf{79.8$\pm$7.2} & \textbf{85.4$\pm$5.8} & \textbf{77.1$\pm$16.4} & \textbf{93.0$\pm$5.1} \\
        \bottomrule
    \end{tabular}
    \vspace{-0.5cm}
\end{table}

\subsection{Effect of Self-Supervision}\label{ss:effect_of_ssl}

We further examined the robustness of each MPM method to sparse or missing input data. We used the same set of trained MPM models and random seeds presented for results in Section~\ref{ss:mpm_results}. However, at test time, we randomly dropped 50\% of the input features for each test sample $x'$. Dropping input features in this experiment translates into setting input pixels to no signal or average values after preprocessing as appropriate (e.g., 0 or median feature value). Test samples in this experiment are considered sparse as only 50\% of the original input features remain. The experimental procedure followed that presented in Section~\ref{ss:experimental_setup}.

Quantitative results for this ablation showed our approach was robust to test sample sparsity while other baseline methods severely degraded in prediction performance, summarized by Table~\ref{tab:robust_abs} (complete results in Appendix~\ref{apdx:robustness}). We believe this sparsity robustness stems from the masked image modeling that occurs during pretraining in our approach. From Figure~\ref{fig:ssl_ex}, it is evident that the features extracted by the self-supervised encoder $\mathcal{E}$ must capture a complete summary of the local geospatial information to enable successful reconstruction. We believe this capability of our approach is especially helpful for MPM in situations where sparsity may be more prevalent (e.g., greenfield exploration).

\subsection{Effect of Likely Negative Sampling}\label{ss:effect_of_neg_sampling}

We also ran experiments comparing our undersampling approach (Section~\ref{sss:pu_learn}) to random undersampling. Random Undersampling (RUS) is common practice within MPM to balance the majority class (i.e., absent or negative labels) with the minority (i.e., present or positive labels). However, it increases the risks of labeling unknown true positives as negatives. Our undersampling method reduces such risks by removing the subset of unknown samples that are most similar to the known positive samples from the undersampling procedure. Empirically, we found it was effective to filter 5-10\% of unknown samples which are most similar to the positive samples. Below we provide some of the quantitative and qualitative results from our experiments, for determining an effective filtering range.

\begin{figure}[t]
  \centering
  \includegraphics[width=0.98\linewidth]{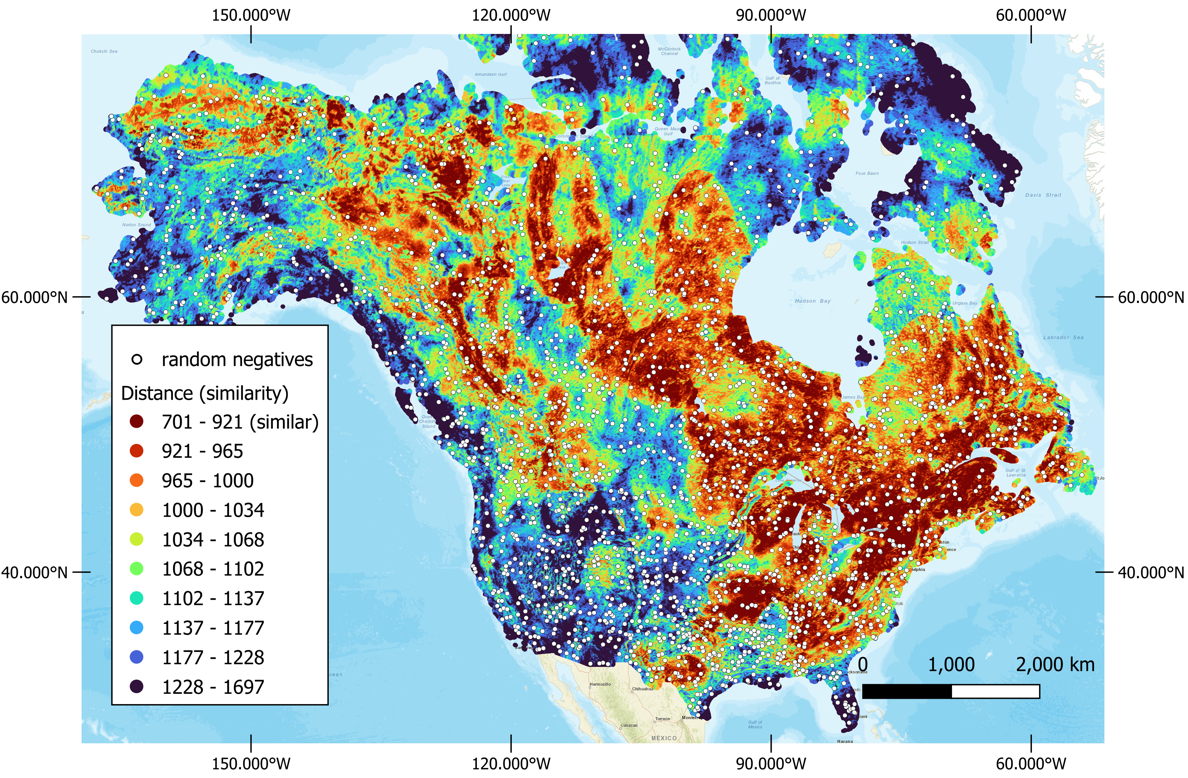}
  \vspace{-0.25cm}
  \caption{Map of samples $x'$ in North America colored by similarity in feature space to the set of known MVT deposits.}
  \label{fig:lnr_abs_rus}
  \Description{Description goes here; less than 2000 characters long (including spaces)}
  \vspace{-0.25cm}
\end{figure}

In Figure~\ref{fig:lnr_abs_rus}, we demonstrate why RUS in prior MPM is problematic. The map plots every sample in North America, as a point with a color indicating the similarity in feature space to the set of known MVT deposits. Navy blue samples are least similar in feature space to available deposits, while burgundy samples are most similar. Note that the burgundy areas overlap well, although in a less discriminative manner, with prospective areas produced in this work, such as Figure~\ref{fig:mvt_midcont_mae}, and prospective areas determined in~\cite{LAWLEY2022104635}. In addition to the samples colored by similarity, we use white points to highlight the negative samples labeled from RUS, as is typical in prior MPM methods. Many of these negative samples are in regions considered prospective, like parts of Missouri and Appalachia in the United States or British Columbia and Nova Scotia in Canada~\cite{LAWLEY2022104635}. Our undersampling approach offers more control over RUS, effectively controlling the range of colors (i.e., similarity in feature space) to choose negative samples.

\begin{figure}[t]
  \centering
  \includegraphics[width=0.98\linewidth]{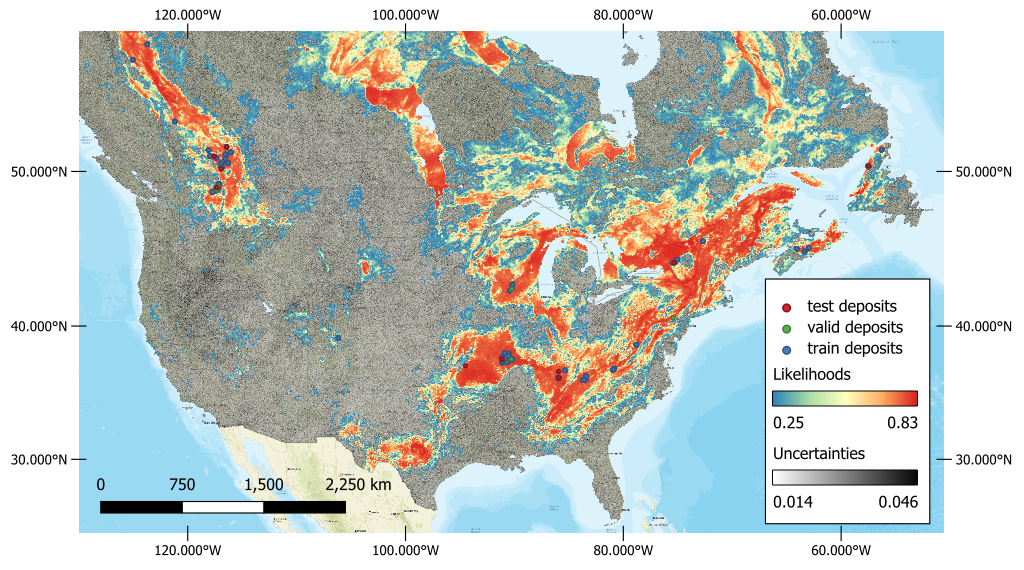}
  \vspace{-0.25cm}
  \caption{Overly generous prospectivity map resulting from training with ``easy'' negatives only (last row in Table~\ref{tab:lnr_abs}).}
  \label{fig:lnr_abs_75}
  \Description{Description goes here; less than 2000 characters long (including spaces)}
\end{figure}

We compared our undersampling approach with RUS, by repeating the MVT deposit assessment with different undersampling settings. We only considered our MPM model in this ablation, as the independent variable considered was the undersampling setting. Table~\ref{tab:lnr_abs} shows how prediction performance is affected by different ranges to filter unknown samples from becoming negatives. In the first row, the broad range filters no unknown samples, which is equivalent to RUS. In the middle row, we have filtered the 10\% of unknown samples which are most similar in feature space to available deposits. These 10\% of unknown samples are no longer available for selection as negative samples. A more extreme case is shown in the final row, where 75\% of unknown samples most similar to available deposits have been filtered. While the metrics may indicate improved performance from filtering a broader range of unknown samples, it is worth considering the prospectivity maps produced from these settings. Figure~\ref{fig:lnr_abs_75} shows the prospectivity maps generated when using the setting in the final row. These maps are clearly overly prospective, stemming from the models training dataset including only ``easy'' negatives. Hence, the final row shows the best prediction performance metrics. Empirically, we observed filtering the 5-10\% of most similar unknown samples balances between introducing noisy negative labels (i.e., RUS) and producing prospectivity maps the are too generous. Appendix~\ref{apdx:lnr_abs} provides additional maps with discussion, for this ablation study.

\section{Conclusion and Future Works}\label{s:conclusion}

MPM seeks to estimate the prospectivity of an exploration target, such as a mineral deposit, at new locations based on knowledge from existing locations. However, MPM is a challenging ML problem due to the extreme rarity of mineralization and complex multi-modal relations that contribute to the mineralization process. While DL holds high potential for MPM, label scarcity within MPM is a limiting factor to gathering sufficient data to train state-of-the-art DL architectures. Additionally, the complexity of DL methods may obfuscate the influence of input features on predictions. 

To address these challenges, we developed a geospatial pretraining approach using masked image modeling that better leverages the unlabeled large-scale multi-modal explanatory features. In our methodology, the pretrained network backbone learns a robust set of features to summarize the local geospatial information, while a small classifier discriminates the prospectivity of those features. We validated our approach by performing mineral assessments in North America and Australia of two deposit types. We compared the outputs of our approach to five existing MPM methods. We observed that our results consistently improved prediction performance over the baselines. Our ablation studies also showed that self-supervision provided model features that were robust to severe input sparsity. Additionally, we used explainable AI techniques to compute feature importances that helped geologically validate individual predictions from our trained DL MPM models. In summary, our findings indicate that self-supervision produces more robust and generalizable features in our DL MPM method.

We view this work as a first step towards developing geospatial foundation models for mineral prospectivity. Recent research in other domains (e.g., computer vision and natural language processing) hints that foundation models may be a promising direction in geospatial modeling. We propose a new approach along this direction, and our experimental results are encouraging. Note there are still many future works remaining for developing full-fledged geospatial foundation models. While we limited our downstream tasks to deposit prospectivity, foundation models in other domains typically serve as backbones for many tasks including classification, segmentation, and others. Building a geospatial foundation model is particularly challenging due to the inherently interdisciplinary nature of the problem integrating Geosciences with Computer Science. We hope our work spurs interest within the community towards this direction.

\begin{table}[t]
    \caption{Summary of Likely Negative Sampling Results; for all the metrics the higher results, the better}
    \label{tab:lnr_abs}
    \vspace{-0.25cm}
    \begin{tabular}{l|c|c|c|c|c|c}
    \hline
    \multicolumn{1}{c|}{\textbf{\begin{tabular}[c]{@{}c@{}}Filter \\ Range\end{tabular}}} & \textbf{F1} & \textbf{MCC} & \textbf{AUPRC} & \textbf{B.ACC} & \textbf{AUROC} & \multicolumn{1}{c}{\textbf{ACC}} \\ \hline
    0\%                                                                                    & 78.9        & 97.8         & 79.8           & 97.8           & 90.9           & 77.9                              \\
    \hline
    10\%                                                                                   & 88.2        & 91.9         & 88.1           & 98.9           & 91.5           & 87.8                              \\
    \hline
    75\%                                                                                   & 90.0        & 98.9         & 100.0          & 98.9           & 99.4           & 89.9                             \\
    \hline
    \end{tabular}
    \vspace{-0.25cm}
\end{table}

\begin{acks}
    This material is based upon work supported by the Defense Advanced Research Projects Agency (DARPA) under Agreement No. HR00112390130.
\end{acks}





\clearpage
\bibliographystyle{ACM-Reference-Format}
\bibliography{bibliography}

\clearpage

\appendix

\section{Extended Experimental Results}

\subsection{Extended Effect of Self-Supervision Ablations} \label{apdx:robustness}

Tables~\ref{tab:apdx_robust_mvt} and \ref{tab:apdx_robust_cd} provided the complete set of predictions metrics for the ablation study testing each method's robustness to sparse input samples.

\begin{table*}[b]
    \caption{Feature Robustness Results for MVT; for all the metrics the higher results, the better; mean $\pm$ std over 5 random seeds}
    \label{tab:apdx_robust_mvt}
    \begin{tabular}{l|cccccc}
        \toprule
        \textbf{Baseline} & \textbf{F1}           & \textbf{MCC}          & \textbf{AUPRC}        & \textbf{B.ACC}        & \textbf{AUROC}$^\dagger$        & \textbf{ACC}$^\dagger$          \\
        \midrule
        WoE               & 6.5$\pm$5.5           & 3.4$\pm$5.4           & 8.0$\pm$1.5           & 51.6$\pm$2.2          & 55.8$\pm$6.2          & 92.2$\pm$2.0                 \\
        \hline
        GBM               & 12.4$\pm$12.5         & 16.1$\pm$14.6         & 44.7$\pm$10.5         & 53.7$\pm$4.1          & 89.2$\pm$3.2          & 95.2$\pm$2.0          \\
        \hline
        ANN               & 35.3$\pm$7.2          & 33.2$\pm$6.9          & 37.0$\pm$4.5          & 68.0$\pm$7.3          & 87.7$\pm$1.7          & 93.2$\pm$1.4          \\
        \hline
        CNN               & 47.9$\pm$6.2          & 46.1$\pm$5.8          & 44.5$\pm$8.7          & 75.2$\pm$4.3          & 87.3$\pm$2.0          & 93.9$\pm$2.2          \\
        \hline
        ViT               & 54.9$\pm$11.9         & 53.1$\pm$12.2         & 64.8$\pm$6.5          & 75.2$\pm$8.3          & 95.2$\pm$2.6          & 95.9$\pm$0.9          \\
        \hline
        Ours              & \textbf{79.8$\pm$7.2} & \textbf{79.2$\pm$7.5} & \textbf{85.4$\pm$5.8} & \textbf{87.9$\pm$2.7} & \textbf{96.7$\pm$2.9} & \textbf{98.0$\pm$0.9} \\
        \bottomrule
        \multicolumn{7}{r}{\footnotesize{$\dagger$ Metrics not intended for imbalanced datasets.}} \\ 
    \end{tabular}
    \vspace{-0.25cm}
\end{table*}

\begin{table*}[b]
  \caption{Feature Robustness Results for CD; for all the metrics the higher results, the better; mean $\pm$ std over 5 random seeds}
  \label{tab:apdx_robust_cd}
  \begin{tabular}{l|cccccc}
    \toprule
    \textbf{Baseline} & \textbf{F1}         & \textbf{MCC}          & \textbf{AUPRC}        & \textbf{B.ACC}        & \textbf{AUROC}$^\dagger$        & \textbf{ACC}$^\dagger$      \\
    \midrule
    WoE               & 3.6$\pm$6.2         & -0.7$\pm$5.8          & 9.7$\pm$3.5           & 49.8$\pm$2.7          & 58.6$\pm$13.4         & 92.3$\pm$2.2 \\
    \hline
    GBM               & 15.3$\pm$20.2       & 13.8$\pm$19.2         & 32.5$\pm$16.3         & 56.9$\pm$9.4          & 80.1$\pm$6.1          & 94.6$\pm$0.7      \\
    \hline
    ANN               & 40.7$\pm$6.4        & 40.7$\pm$6.3          & 45.9$\pm$6.2          & 80.3$\pm$4.4          & 89.8$\pm$3.6          & 89.5$\pm$3.5      \\
    \hline
    CNN               & 36.2$\pm$12.2       & 33.0$\pm$13.2         & 38.1$\pm$10.6         & 67.3$\pm$3.9          & 78.1$\pm$9.6          & 92.0$\pm$4.0      \\
    \hline
    ViT               & 62.4$\pm$11.8       & 62.5$\pm$12.0         & 64.9$\pm$15.9         & 80.2$\pm$6.5          & 91.3$\pm$7.5          & 96.2$\pm$1.5      \\
    \hline
    Ours              & \bf{77.1$\pm$16.4}  & \bf{77.5$\pm$15.6}    & \bf{93.0$\pm$5.1}     & \bf{92.4$\pm$5.7}     & \bf{99.3$\pm$0.6}     & \bf{96.8$\pm$3.4}     \\
    \bottomrule
    \multicolumn{7}{r}{\footnotesize{$\dagger$ Metrics not intended for imbalanced datasets.}} \\ 
  \end{tabular}
  \vspace{-0.25cm}
\end{table*}

\subsection{Extended Effect of Likely Negative Sampling} \label{apdx:lnr_abs}

\begin{figure}[b]
  \centering
  \includegraphics[width=0.98\linewidth]{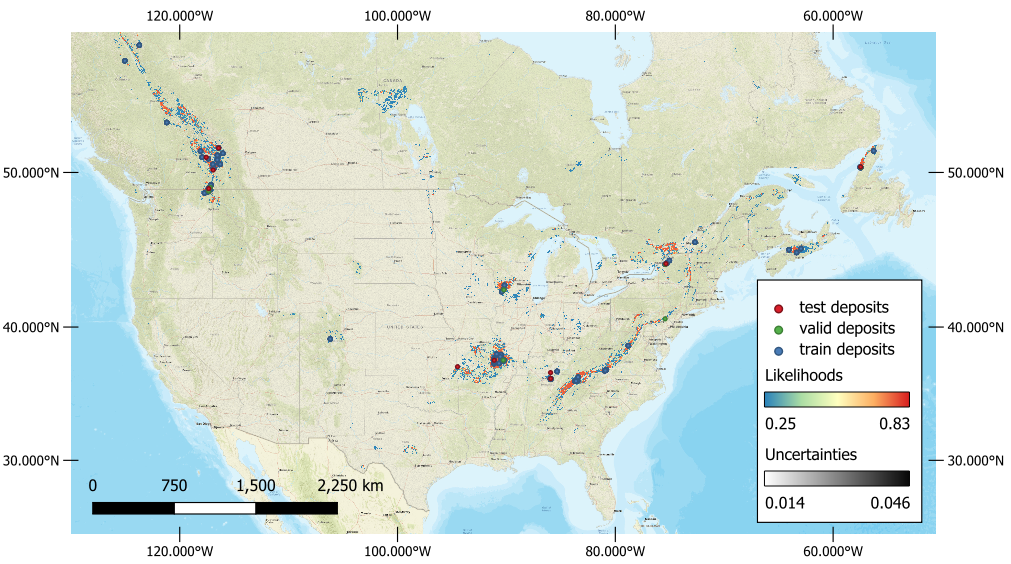}
  \vspace{-0.25cm}
  \caption{Prospectivity map resulting from training with negatives produced using settnig in first row of Table~\ref{tab:robust_abs}.}
  \label{fig:lnr_abs_0}
  \Description{Description goes here; less than 2000 characters long (including spaces)}
\end{figure}

\begin{figure}[b]
  \centering
  \includegraphics[width=0.98\linewidth]{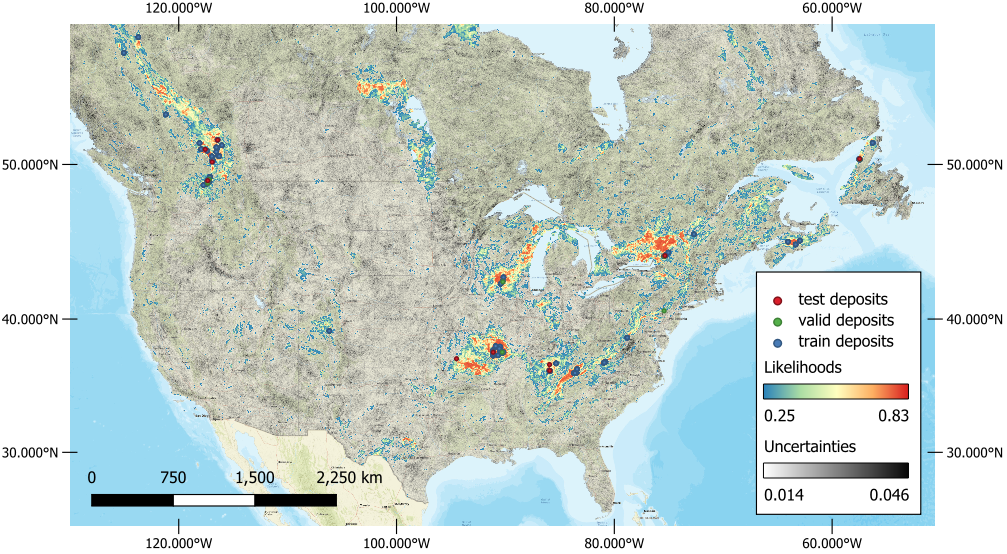}
  \vspace{-0.25cm}
  \caption{Prospectivity map resulting from training with negatives produced using settnig in second row of Table~\ref{tab:robust_abs}.}
  \label{fig:lnr_abs_10}
  \Description{Description goes here; less than 2000 characters long (including spaces)}
\end{figure}

\begin{figure}[b]
  \centering
  \includegraphics[width=0.98\linewidth]{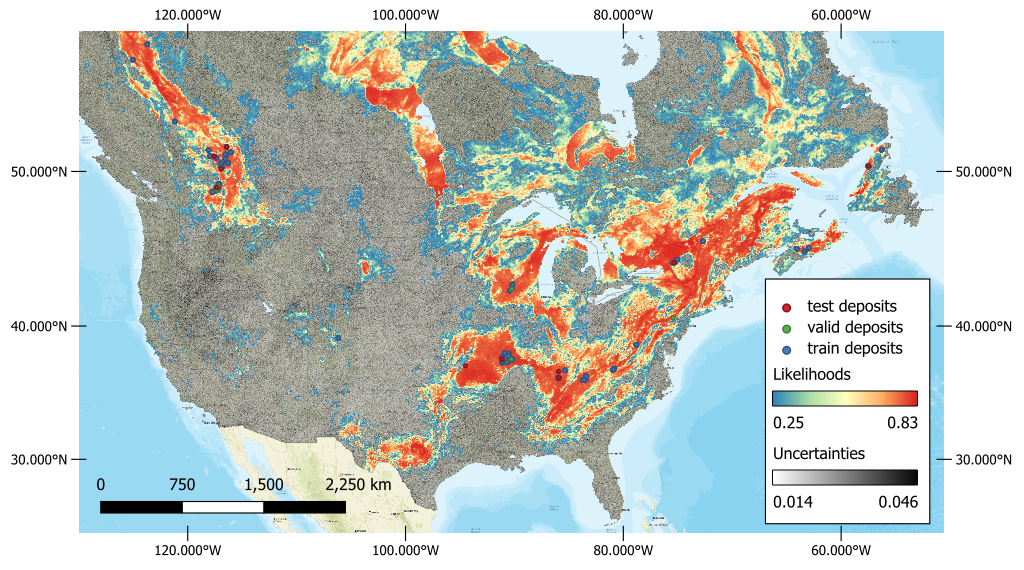}
  \vspace{-0.25cm}
  \caption{Prospectivity map resulting from training with negatives produced using settnig in third row of Table~\ref{tab:robust_abs}.}
  \label{fig:lnr_abs_75_apdx}
  \Description{Description goes here; less than 2000 characters long (including spaces)}
\end{figure}

Figures~\ref{fig:lnr_abs_0} through \ref{fig:lnr_abs_75_apdx} show the prospectivity map produced when using the each setting for filtering unknown samples in Table~\ref{tab:robust_abs} from Section~\ref{ss:effect_of_neg_sampling}

\end{document}